\documentclass{article}

\usepackage{iclr2026_conference,times}

\usepackage[utf8]{inputenc} 
\usepackage[T1]{fontenc}    
\usepackage{hyperref}       
\usepackage{url}            
\usepackage{booktabs}       
\usepackage{amsfonts}       
\usepackage{nicefrac}       
\usepackage{microtype}      
\usepackage{xcolor}         
\usepackage{graphicx}

\usepackage{amsmath, amssymb,bm}
\usepackage{hyperref}
\usepackage{mathtools}
\usepackage{amsthm}
\usepackage{cite}
\usepackage{microtype}
\usepackage{multirow}
\usepackage{enumitem}
\usepackage{mathrsfs}
\usepackage{xcolor}
\usepackage{tikz}
\usepackage{pgfplots}
\usepackage{subcaption}

\definecolor{copper}{HTML}{C84E00}
\definecolor{dandelion}{HTML}{FFD960}
\definecolor{piedmont}{HTML}{A1B70D}
\definecolor{eno}{HTML}{339898}
\definecolor{shale}{HTML}{0577B1}
\definecolor{dukeblue}{HTML}{012169}
\definecolor{ironweed}{HTML}{993399}
\definecolor{whisper}{HTML}{F3F2F1}
\definecolor{gingerbeer}{HTML}{FCF7E5}

\hypersetup{
  colorlinks=true,
  linkcolor=dukeblue!70!white,
  urlcolor=dukeblue!70!white,
  citecolor=dukeblue!70!white
}

\newtheorem{theorem}{Theorem}

\newtheorem{lemma}{Lemma}

\theoremstyle{definition}

\newcommand{\bC}{\bm{C}}

\newcommand{\bG}{\bm{G}}

\newcommand{\bM}{\bm{M}}

\newcommand{\bP}{\bm{P}}
\newcommand{\bQ}{\bm{Q}}
\newcommand{\bR}{\bm{R}}
\newcommand{\bS}{\bm{S}}

\newcommand{\bX}{\bm{X}}

\newcommand{\bw}{\bm{w}}

\newcommand{\by}{\bm{y}}

\newcommand{\bbR}{\mathbb{R}}

\newcommand{\one}{\boldsymbol{1}} 
\newcommand{\Id}{\mathrm{I}}

\newcommand{\btheta}{\bm{\theta}}

\DeclareMathOperator{\gvec}{\mathsf{vec}}

\DeclareMathOperator{\diag}{\mathrm diag}

\newcommand{\eps}{\epsilon}
\newcommand{\normal}{\mathsf{N}}

\newcommand{\E}{\mathbb{E}}
\renewcommand{\P}{\mathbb{P}}

\DeclarePairedDelimiter\bkt{[}{]}     
\makeatletter
\newcommand{\@exstar}[1]{\E \bkt*{#1}}
\newcommand{\@exnostar}[2][]{\E \bkt[#1]{#2}}
\newcommand{\ex}{\@ifstar\@exstar\@exnostar}
\makeatother

\makeatletter
\newcommand{\@prstar}[1]{\P \bkt*{#1}}
\newcommand{\@prnostar}[2][]{\P \bkt[#1]{#2}}
\newcommand{\pr}{\@ifstar\@prstar\@prnostar}
\makeatother

\DeclareMathOperator{\cov}{\mathsf{Cov}}

\pgfplotsset{compat=newest} 

\let\originalleft\left
\let\originalright\right
\renewcommand{\left}{\mathopen{}\mathclose\bgroup\originalleft}
\renewcommand{\right}{\aftergroup\egroup\originalright}

\mathtoolsset{showonlyrefs}

\newcommand{\ejw}[1]{\textcolor{black}{#1}}

\newcommand*\halfcirc[1][1ex]{%
  \begin{tikzpicture}
  \draw[fill] (0,0)-- (90:#1) arc (90:270:#1) -- cycle ;
  \draw (0,0) circle (#1);
  \end{tikzpicture}}
\newcommand*\fullcirc[1][1ex]{\tikz\fill (0,0) circle (#1);}

\allowdisplaybreaks

\newcommand{\para}[1]{{\vspace{2pt} \bf \noindent #1}}  

\usepackage{tcolorbox}

\newtcolorbox{callout}{
    colback=gingerbeer!80,
    colframe=black,
    coltext=black,
    boxrule=1pt,
    arc=5pt,
    left=5pt,
    right=5pt,
    top=5pt,
    bottom=5pt,
}
\newenvironment{packed_itemize}{
\begin{list}{\labelitemi}{\leftmargin=1em}
\setlength{\itemsep}{1pt}                                                           
\setlength{\parskip}{0pt}                                                                                 \setlength{\parsep}{0pt}                                                                                  \setlength{\headsep}{0pt}                                                                                 \setlength{\topskip}{0pt}                                                                                 \setlength{\topmargin}{0pt}                                                                               \setlength{\topsep}{0pt}                                                                                  \setlength{\partopsep}{0pt}                                                                               }{\end{list}}                                                                                                                         

\newenvironment{packed_itemize_ejw}{
\begin{list}{\labelitemi}{\leftmargin=1em}
\setlength{\itemsep}{1pt}                                                           
\setlength{\parskip}{0pt}                                                                              \setlength{\topskip}{0pt}                                                                       \setlength{\topmargin}{0pt}                                                                                         \setlength{\topsep}{0pt}                                                                                  \setlength{\partopsep}{0pt}                                                                               }{\end{list}} 
\newenvironment{packed_enumerate}{  
\begin{enumerate}[leftmargin=2em]                  \setlength{\itemsep}{1pt}                                        \setlength{\parskip}{0pt}                                         \setlength{\parsep}{0pt}                                  
    \setlength{\headsep}{0pt}                                                                                 \setlength{\topskip}{0pt}                                                                                 \setlength{\topmargin}{0pt}                                                                               \setlength{\topsep}{0pt}                                                                                  \setlength{\partopsep}{0pt}
}{\end{enumerate}}

\newcommand{\neutral}[1]{\colorbox{dandelion!50!white}{#1}}
\newcommand{\decrease}[1]{\colorbox{piedmont!50!white}{#1}}
\newcommand{\increase}[1]{\colorbox{copper!50!white}{#1}}

\title{What happens when generative AI models train recursively on each others' outputs?}

\author{
    Hung Anh Vu, Galen Reeves, \& Emily Wenger\thanks{Corresponding author: \texttt{emily.wenger@duke.edu}} \\
    Department of Electrical and Computer Engineering\\
    Duke University
}

\iclrfinalcopy
\begin{document}

\maketitle

\begin{abstract}
The internet serves as a common source of training data for generative AI (genAI) models but is increasingly populated with AI-generated content. This duality raises the possibility that future genAI models may be trained on other models'  generated outputs. Prior work has studied consequences of models training on their own generated outputs, but limited work has considered what happens if models ingest content produced by other models. Given society's increasing dependence on genAI tools, understanding such data-mediated model interactions is critical. This work provides empirical evidence for how data-mediated interactions might unfold in practice, develops a theoretical model for this interactive training process, and experimentally validates the theory. We find that data-mediated interactions can benefit models by exposing them to novel concepts perhaps missed in original training data, but also can homogenize their performance on shared tasks.
\end{abstract}

\section{Introduction}
\vspace{-.2cm}
Since the release of ChatGPT in 2022, generative AI (genAI) models have exploded in popularity. Now capable of generating highly realistic text, images, and videos, these models have been widely adopted for various use cases, from creative idea generation~\citep{notion_ad, grammarly_ad} to healthcare support~\citep{reddy2024generative} to national security settings~\citep{dod_ai_use_1, health_ai_1}. Given the significant uptick in genAI use across numerous industries, this technology is clearly here to stay. Consequently, interrogating potential ways genAI models could evolve\textemdash in positive or harmful ways\textemdash is critical. 

With few exceptions, today's large-scale genAI models are trained on massive datasets sourced from the internet. Widely-accepted scaling laws for model performance say that training on more data aids learning~\citep{kaplan2020scaling}, and the internet provides a rich, cheap, and ever-evolving source of training data. Although whitepapers for more recent genAI models withhold details about training set composition\textemdash potentially due to ongoing litigation about copyright concerns\textemdash evidence from earlier whitepapers indicates that scraped data was used to train models like Llama, Gemini, Phi, the GPT series, Claude, and others~\citep{dubey2024llama, achiam2023gpt, team2024gemini, jiang2023mistral, cohere, claude2, abdin2024phi}. 

Beyond privacy and copyright concerns, training on scraped data could have other downsides. Prior work has noted that genAI models trained recursively on their own generated outputs ``collapse,'' becoming unable to generate meaningful content~\citep{shumailov2024ai, hataya2023will, martinez2023towards,  alemohammad2023self}. This scenario is feasible, since AI-generated content abounds online~\citep{sun2024we} and could be part of future training datasets. However, subsequent work has proposed ways to mitigate collapse via reuse of non-AI-generated data in subsequent training iterations~\citep{dey2024universality, kazdan2024collapse, dohmatob2025model, feng2024beyond}. Model collapse remains an activate research area~\citep{schaeffer2025position}.  

Yet, prior work studying the dynamics of model collapse has overlooked another reality: the internet teems with content from {\em many} genAI  models. Today's most popular models have millions of users~\citep{reuters_openai, handa2025economic, llama_users}, who leverage generative AI tools to create online content like web pages and social media posts~\citep{genai_creative}. Recent work showed that up to 40\% of content on popular sites like Quora is now AI-generated~\citep{sun2024we}. Given the increasing availability of these models for a variety of public-facing uses~\citep{apple_ad, notion_ad, grammarly_ad}, AI-generated content from many different models will continue to proliferate.

The standard practice of training on scraped internet data and the increasing prevalence of AI-generated content online suggest the strong possibility that {\em future generative AI models will be trained on other models' outputs.} Yet, this aspect of model training has received relatively little attention. Given the widespread adoption of generative AI models in critical settings like healthcare and national security, this phenomenon ought to be investigated to ensure models remain helpful and trustworthy. 

\para{Contributions.} To address this need, this work theoretically derives and experimentally evaluates the long-term evolutionary behavior of generative models trained on {\em each other's} data. Specifically, we
\begin{packed_itemize}
    \item Develop a framework, grounded in real-world evidence, describing data-mediated interactions between genAI models. 
    \item Derive concise formulas describing the dynamics of interactive training under varied regimes.
    \item Run experiments on large language models to understand how data-mediated interactions affect model performance in practice.
\end{packed_itemize}

Both our theoretical analysis and experiments show that when training with a mixture of real and synthetic data, the implicit interaction between heterogeneous models and datasets can have both positive and negative impacts. At a high level, well known concepts in statistical learning theory anticipate this: recursive training on the same data is bad (e.g., overfitting and model ``collapse'') but training on novel data, even if synthetic, can boost performance (e.g., transfer learning). Our experimental results provide concrete evidence that these phenomena can occur simultaneously. 
\vspace{-0.2cm}
\vspace{-0.2cm}
\section{Related Work}
\vspace{-0.2cm}

\para{Model collapse} is a recently observed phenomenon in large-scale generative text and image models. It referred\textemdash in it's earliest form\textemdash to the phenomenon of models performing much worse after generations of training on their own generated outputs~\citep{shumailov2024ai,peterson2025ai, wang2024bias,alemohammad2023self, feng2024beyond, hataya2023will, dohmatob2025model, martinez2023towards}. Theoretical and empirical results from these works show that models, if trained on generated outputs from their prior versions, slowly degrade in performance as generations progress. One way this manifests is in models forgetting the tails of their original (real) training data, since generated content tends not to contain rare content from the original training data. Training repeatedly on truncated, synthetic data leads the model to forget the richness of its original distribution, resulting in degraded behavior (at best) and total failure (at worst).

\para{Mitigating model collapse.} Despite the dire predictions of these papers, subsequent work has proposed a simple mitigation strategy: instead of discarding all prior (human-generated) training data, retain some fraction of this while augmenting it with generated data. Numerous works have observed that this choice to {\em augment} instead of {\em discard} the original training dataset results in a bounded error in future models, avoiding collapse~\citep{kazdan2024collapse, gerstgrasser2024model, marchi2024heat}. Although most of these results were discovered on small models, recent work claims that the observed bound in error $\pi^2/6$ exists for all models~\citep{dey2024universality}. Further work~\citep{schaeffer2025position} summarizes current research on collapse. 

\para{Transfer learning and other model interactions.} Significant prior work has studied transfer learning, in which information learned by one model is passed to another, often by reusing the trained weights of a ``teacher'' model to initialize a ``student'' model~\citep{zhuang2020comprehensive}. Some prior work has further considered the use of synthetic data in transfer learning~\citep{tian2025generative, kim2022transferable, brinner2025enhancing}. Our work is distinct from transfer learning due to its focus on {\em unintentional} data-mediated interactions between models. Furthermore, limited work has examined long-term effects of models training on each other's data. ~\citet{zhang2024regurgitative} consider the setting where a generative model is trained on data generated by other models, but does not consider long-term effects of such interactions among multiple models. ~\citet{jain2025interacting} study interacting Large Language Model agents through the lens of Bayesian social learning and microeconomics, but do not focus specifically on data-mediated interactions between models.
\section{How Today's Large-Scale Generative AI Models Are Trained}
\label{sec:realities}
\vspace{-0.2cm}

We first establish {\em why} we believe that data-mediated interactions between models\textemdash e.g. instances of models training on each other's generated outputs\textemdash are realistic and worthy of study. To do this, we comb through academic literature and whitepapers describing today's large-scale genAI models, focusing specifically on large language models, to understand how models are trained, what data they are trained on, and how data is collected and used for model updates. This section sets the stage for the formalization, theory, and experiments in the rest of the paper. 

Our literature review reveals that most of today's models follow a 3 step update process. First, models are {\bf pretrained} on a large corpus of data; then they are {\bf fine-tuned} to teach specific behaviors and/or to align them with human preferences. Finally, they are later {\bf updated}, either to teach new behaviors or update knowledge. As we describe these steps in detail below, we highlight specific realities or assumptions that have been largely overlooked or not made explicit by prior work. 

\para{Step 1: Pretraining.} Following well-established scaling laws linking model performance and dataset size~\citep{kaplan2020scaling}, today's large-scale generative AI models are trained on massive datasets often scraped from the internet. Early versions of GPT, Llama, and PaLM all report being trained on scraped datasets like Common Crawl, ArXiv, Github, Wikipedia, and/or Stack Exchange~\citep{touvron2023llama, chowdhery2023palm, brown2020language}\textemdash see Table~\ref{tab:train-data} for an overview. 
\begin{table*}[t]
\centering
\caption{\small {\bf Examples of training data listed for prominent language models.} \em \fullcirc = use explicitly stated; \halfcirc = significant overlap expected (e.g. GLaM~\citep{du2022glam} and PaLM~\citep{chowdhery2023palm} are trained on Microsoft's internal re-creation of WebText~\citep{brown2020language}). We only include models for which training data sources are explicitly stated. See Table~\ref{tab:training_info} in Appendix for information on other prominent models.}
\label{tab:train-data}
\resizebox{\textwidth}{!}{%
\begin{tabular}{ccccccccc}
\toprule
Model &  CommonCrawl & WebText & Github & Wikipedia & Books & ArXiv & StackExchange & News \\ \midrule
Chinchilla~\citep{hoffmann2022training} & \fullcirc & & \fullcirc & \fullcirc & \fullcirc & & &  \fullcirc \\ 
GLaM~\citep{du2022glam} &  & \halfcirc & & \fullcirc & \fullcirc &  & & \fullcirc \\ 
GPT~\citep{radford2018improving} &  & &  & & \fullcirc & & \\  
GPT-2~\citep{radford2019language}  &  & \fullcirc & & & & & \\ 
GPT-3~\citep{brown2020language}     & \fullcirc & \fullcirc &  & \fullcirc & \fullcirc & & & \fullcirc \\
LaMDA~\citep{thoppilan2022lamda} & \fullcirc & & & \fullcirc & & & \halfcirc & \\
Llama 1~\citep{touvron2023llama}       & \fullcirc & & \fullcirc & \fullcirc & \fullcirc  & \fullcirc  & \fullcirc & \\
PaLM~\citep{chowdhery2023palm}   &  & \halfcirc & & \fullcirc & \fullcirc & & & \fullcirc \\ 
Phi 2~\citep{phi2} & \fullcirc  & & & \fullcirc & \fullcirc &\fullcirc & \fullcirc & \\
\bottomrule

\end{tabular}%
}

\vspace{-0.2cm}
\end{table*}


\begin{callout}
    \centering {\bf Reality}: large-scale genAI models are pretrained on internet-scraped datasets. 
\end{callout}

Another striking fact emerges from the categorization of training data in Table~\ref{tab:train-data}: large-scale model training datasets overlap. For example, GPT, Jamba, Llama, PaLM, and Phi are all trained on subsets of CommonCrawl~\citep{commoncrawl}, while GPT, Llama, and PaLM are all trained on Wikipedia and Books datasets. Several other models have other points of training data overlap. This suggests that:

\begin{callout}
    \centering {\bf (Previously overlooked) reality:} internet-scraped AI training datasets overlap substantially. 
\end{callout}


\begin{table*}[t]
\centering
\caption{\small {\bf Evidence from LLama, GPT, and Phi suggests reuse of old training and collection of additional data to train new model generations.} \em Datasets reused across models and generations are highlighted. We start with Phi 1.5, the first version of Phi designed for general NLP tasks. Phi 1 was designed for coding tasks.}
\vspace{-0.2cm}
\label{tab:generations}
\resizebox{\textwidth}{!}{%
\begin{tabular}{ccccc}
\toprule
\textbf{Model} &
  \textbf{v1} &
  \textbf{v2} &
  \textbf{v3} &
  \textbf{v4} \\ \midrule
Llama &
  \begin{tabular}[c]{@{}c@{}}\citep{touvron2023llama}:\\ ArXiv, \colorbox{dukeblue!30!white}{Books}, \colorbox{dandelion!50!white}{Common Crawl},\\ C4, Wikipedia, StackExchange\end{tabular} &
  \begin{tabular}[c]{@{}c@{}}\citep{touvron2023llama_b}:\\"A new mix of publicly\\available online data."\end{tabular} &
  \begin{tabular}[c]{@{}c@{}}\citep{dubey2024llama}:\\ "A variety of data sources containing\\ knowledge until the end of 2023."\end{tabular} &
   \begin{tabular}[c]{@{}c@{}}\citep{llama4}:\\ "A mix of publicly available,\\licensed data and information from\\Meta’s products and services."\end{tabular}\\ \midrule
GPT &
 \begin{tabular}[c]{@{}c@{}} \citep{radford2018improving}: \\ \colorbox{dukeblue!30!white}{BooksCorpus}\end{tabular} &
\begin{tabular}[c]{@{}c@{}}  \citep{radford2019language}: \\ \colorbox{eno!40!white}{WebText} \end{tabular} &
  \begin{tabular}[c]{@{}c@{}}\citep{brown2020language}:\\ \colorbox{dandelion!50!white}{CommonCrawl}, \colorbox{eno!40!white}{WebText2},\\ \colorbox{dukeblue!30!white}{Books, Books2}, Wikipedia\end{tabular} &
  \begin{tabular}[c]{@{}c@{}}\citep{achiam2023gpt}: No info\\ provided\end{tabular} \\ \midrule
  Phi & \begin{tabular}[c]{@{}c@{}}\citep{li2023textbooks}:\\ \colorbox{copper!40!white}{The Stack}, \colorbox{ironweed!40!white}{Stack Overflow},\\\colorbox{whisper!95!black}{synthetic ``textbook'' data}\end{tabular} & \begin{tabular}[c]{@{}c@{}} \citep{phi2}:\\ \colorbox{copper!40!white}{The Stack}, \colorbox{ironweed!40!white}{Stack Overflow},\\ \colorbox{whisper!95!black}{synthetic ``textbook'' data,}\\ \colorbox{dandelion!50!white}{filtered Commmon Crawl}\end{tabular} & \begin{tabular}[c]{@{}c@{}} \citep{abdin2024phi}:\\ "publicly available web data\ldots \\and synthetic LLM-generated data" \end{tabular} & N/A \\ \bottomrule
\end{tabular}%
}
\end{table*}

\para{Step 2: Fine-tuning.} Variously called fine-tuning or alignment, this phase leverages proprietary methods or data to tweak model behaviors in ways model providers believe are helpful. For example, the LLama fine-tuning phase~\citep{dubey2024llama} involves many rounds of reinforcement learning with human feedback (RLHF) to stamp out model negative behaviors, while Phi~\citep{abdin2024phi} was fine-tuned on proprietary synthetic data to patch ``gaps'' in its mathematical reasoning abilities. 
\begin{callout}
    \centering    {\bf Reality:} large-scale genAI models are (typically) fine-tuned using proprietary datasets/methods.
\end{callout}

\para{Step 3: Model updates.} A key assumption of prior literature on model collapse is that models are {\em updated}.  Updating generally involves training a new model using similar architecture initialized with weights from the prior step, techniques, and datasets to prior versions of the model. This matches reality\textemdash the literature documents numerous ``families'' of models that directly descend from one another (c.f. LLama 1, 2, and 3~\citep{touvron2023llama, touvron2023llama_b, dubey2024llama}; GPT 1, 2, 3, 4~\citep{radford2018improving, brown2020language, radford2019language, achiam2023gpt}; Phi 1.5, 2, 3~\citep{li2023textbooks, phi2, abdin2024phi}). Prior works on model collapse have proposed several ways training data from prior model generations could be re-used (or not) during model updates.

The {\em replace} scenario, first proposed by~\citet{shumailov2024ai} assumes model trainers completely replace their training data at each update, using {\em only generated data output by the prior version of the model.}~\citet{schaeffer2025position} point out that this edge case is interesting but impractical. The {\em accumulate} scenario~\citep{gerstgrasser2024model} assumes model trainers augment their original data with additional data that may contain AI-generated content at each update set. Finally, the {\em accumulate and subsample} scenario~\citep{kazdan2024collapse} echoes the accumulate scenario but subsamples a fixed-sized subset from the original and accumulated data at each update step (rather than letting dataset size grow linearly). This scenario acknowledges the real-world compute limitations. 

{\em \bf We believe that the accumulate-and-subsample paradigm best reflects reality,} so we will leverage it in our work. We support this opinion with evidence from three well-documented model families: Llama, GPT, and Phi. Table~\ref{tab:generations} records the training data used in publicly disclosed generations of these models. As the table shows, trainers re-use some prior training data for model updates, supplementing this with additional web content. Whitepapers for models published after 2023 do not contain specific training data information but suggest collection of additional online data at each update.

\begin{callout}
    {\bf Reality:} model trainers reuse data from prior generations and supplement this with new (internet-scraped) content when updating models.
\end{callout}

However, most prior work on model collapse still overlooks a fundamental reality in model updates: {\em future models trained on internet-sourced content will be trained on outputs from other generative AI models}, not merely their own. Already, the internet is filled with generated content from various models~\citep{sun2024we}. As model trainers collect new data to facilitate model updates, internet-sourced data will inevitably contain content from other generative models. 

\begin{callout}
    {\bf (Previously overlooked) reality:} at each update step, models may be trained on {\em their own} and {\em other models'} generated outputs.
\end{callout}

\vspace{-0.2cm}
\section{Formalizing An Iterative, Interactive Model Training Pipeline}
\label{sec:formal}
\vspace{-0.2cm}

\begin{figure*}[t]
    \centering
    \includegraphics[width=0.85\textwidth]{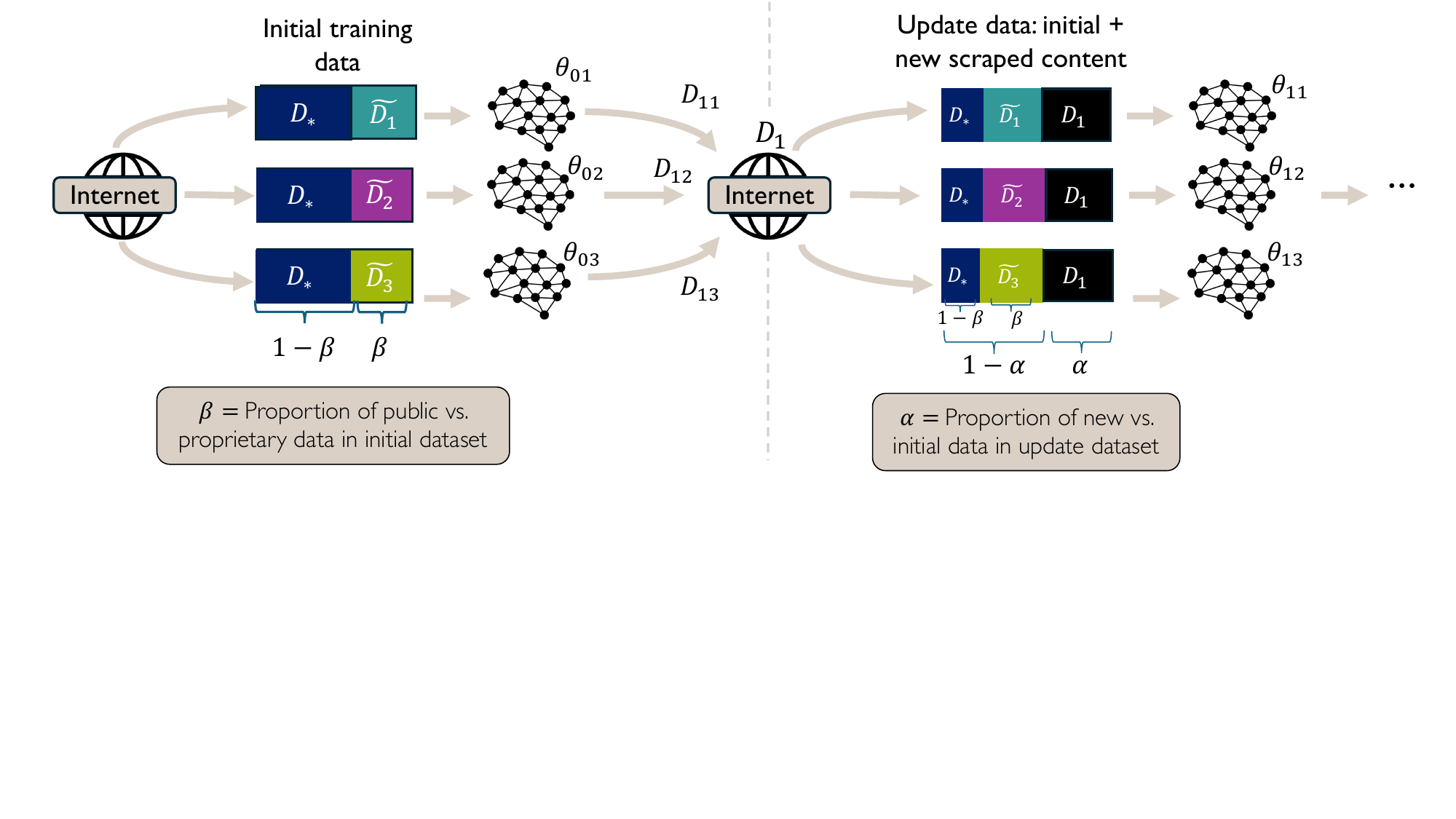}
    \vspace{-3.1cm}
    \caption{\small {\bf Our dataset update scheme, parameterized by $\alpha$ and $\beta$.} \em This paradigm best aligns with evidence from the literature given in \S\ref{sec:realities} and strongly indicates that interactions between models, facilitating by training on others' generated data, are an important consideration for empirical and theoretical work on model evolution. }
    \label{Fig: Interaction setup}
    \vspace{-0.3cm}
\end{figure*}

Section~\ref{sec:realities} provides empirical evidence for two realities overlooked by prior studies: internet-scraped training datasets used for initial training may have substantial overlap, and models may be updated using {\em each others' generated outputs}. 
To study the effect of these two factors on model evolution, we propose a general workflow in which multiple entities regularly update their models using a mix of private, public, and generated data. Based on \S\ref{sec:realities}, we consider three types of training/update data:
\begin{packed_itemize}
    \item $D_*$: Public data used during initial training/updates by multiple entities (real data only).
    \item $\tilde{D}_k$: Private data used only by entity $k$ for initial training/updates (real data only).
    \item $D_{t} = \{D_{t1}, D_{t2}, \ldots D_{tk}\}$: Public data used for updates at time $t$ by multiple entities (synthetic data). $D_{tk}$ is data generated by the $k^{th}$ entity based on model $\theta_{t-1,k}$.
\end{packed_itemize}
Mapping these to realistic scenarios, $D_{*}$ could be a public dataset like Common Crawl; $\tilde{D}_k$ could be a private dataset of math problems curated by entity $k$; and $D_{t}$ could be an internet scrape from after initial model training. We weight the relative impact of these data types by the ratios $\alpha, \beta$. 
\begin{packed_itemize_ejw}
   \item $\beta$, $0 \le \beta \le 1$, is relative size of the initial public data set $D_{*}$ compared to the initial private data set $\tilde{D}_k$. This fraction remains constant if/when initial data is reused for updates. 
    \item $\alpha$, $0 \le \alpha \le 1$ is the fraction of new data introduced at generation $t$, relative to the amount of initial data reused (following the ``accumulate and subsample'' paradigm of~\citet{kazdan2024collapse}).  
\end{packed_itemize_ejw}

\para{Interactive training workflow.} We consider $K$ entities, each seeking to train or update their own generative AI model. In the initial phase of training, denoted by time $t = 0$, each entity $k$ trains its model based on a combination of a publicly available dataset $D_{*}$ as well its own private dataset $\tilde{D}_k$. The trained model is represented generally by a parameter $\hat{\theta}_{t,k}$, i.e., 
$\hat{\theta}_{0,k} = \Phi_{0,k}( D_{*} ,\tilde{D}_k)$, $k = 1, \dots, K$.
where each $\Phi_{0,k}$ represents a generic training algorithm.
For model updates at stages $t > 0$, model parameters are updated via: 
\begin{packed_enumerate}
\item New public data $D_{t}$ is generated uniformly at random using the most recent version of the models. Specifically, the data are sampled i.i.d.\ according to the mixture 
$\frac{1}{K} \sum_{k=1}^K P_{k,\hat{\theta}_{t-1,k}}$ 
where $P_{k,\theta}$ denotes the generative model used by $k$-th entity.  
\item This data is placed online and collected by entities as training data for the next model update. 
\item Each entity composes its training data for the next update, using a mix of the initial dataset $(D_*, \tilde{D}_k)$ and newly collected data $D_t$. Contributions from each dataset are weighted by $\alpha, \beta$.

\item Each entity $k = 1, \dots, K$ updates it model parameters via
$\hat{\theta}_{t,k} = \Phi_{t,k}(\hat{\theta}_{t-1,k}, D_{*} , \tilde{D}_k, D_t)$.
Here, $\Phi_{0,k}$ is a training algorithm that depends on the previous model parameter $\hat{\theta}_{k,t-1}$ as well as the data. As before, training may employ subsampling, weighting, and randomization. 
\end{packed_enumerate} 

In this workflow, entities interact through the release of publicly available synthetic data produced by prior generations of other entities' models. Thus, even though initial private training data are never shared, it could end up positively impacting other entities' models.  This potential benefit of synthetic data sharing appears only in this interaction paradigm and has not be recognized in prior work. 

\vspace{-0.2cm}
\section{Theory}\label{sec:theory}
\vspace{-.3cm}


We theoretically analyze the behavior of the interactive workflow. Similar to prior work \citep{gerstgrasser2024model, kazdan2024collapse, dey2024universality, dohmatob2025model, barzilai2025models}, we focus on the linear regression models where each data point consists of a feature-response pair $(x, y) \in \bbR^d \times \bbR$. By the universality results of  \citet{dey2024universality}, the analysis of this settings also applies to generalized linear models satisfying appropriate asymptotic normality assumptions. 

\para{Notation.} For a $p \times q$ matrix $A$, we use $A^+$ to denote the Moore-Penrose pseudoinverse and $\gvec(A)$ to denote the $pq \times 1$ vector obtained by stacking the columns. $\otimes$ denotes the Kronecker product. 
For $\alpha, \beta \in [0,1]$ we set $\bar{\alpha} = 1- \alpha$ and $\bar{\beta} = 1- \beta$. 

\para{Training Workflow.} We follow the training pipeline outlined in Section~\ref{sec:formal} in which $K$ different models are trained on mixture of private, public, and generated data. 
At initialization,  each entity $k \in [K]$ combines its private data $\tilde{D}_k = (\tilde{x}_{ki}, \tilde{y}_{ki})_{i=1}^{\tilde{n}_k}$ with public data $D_* = (x_{*i}, y_{*i})_{i=1}^{n_*}$ to produce an  estimate $\hat{\theta}_{k0}$ by minimizing the empirical loss 
\begin{align}
\textstyle  \sum_{(x,y)  \in \tilde{D}_k } \beta_0  L(x,y, \theta)  +   \sum_{(x,y)  \in \tilde{D}_* } \bar{\beta}_0 L(x,y, \theta) 
\end{align}
where $L(x,y, \theta) \coloneqq (y - x^\top \theta)^2$ is the squared error loss and $0 \le \beta_0 \le 1$ controls the relative weight placed on the private data. Training then proceeds for generation stages $t = 1,2,3,\dots$ as follows:
\begin{packed_enumerate}
\item  Each entity $k$ uses its most recent parameter estimate $\hat{\theta}_{t-1,t}$ to  generate new data $D_{tk} = (x_{tki}, y_{tki})_{i=1}^{n_{tk}}$ according to the Gaussian model 
$y \mid x \sim \normal( x^\top \theta_{t-1,k} , \sigma^2)$. The entire collection of generated samples is combined into a single public data set $D_t = \cup_{k=1}^K D_{tk} $. 

    \item Each entity $k$ produces a new estimate $\hat{\theta}_{tk}$ by minimizing the empirical loss
        \begin{align}
\textstyle  \sum_{(x,y) \in D_k}    \bar{\alpha}_t \beta_t   L(x,y, \theta)   + \sum_{(x,y) \in D_*}   \bar{\alpha}_t \bar{\beta}_t  L(x,y, \theta) +    \sum_{(x,y) \in D_t}  \frac{\alpha_t}{K} L(x,y, \theta)
    \end{align}
 with weights $0 \le \alpha_t, \beta_t \le 1$. 
 \end{packed_enumerate}

\ejw{We note that our framework could easily be extended to accommodate new, human-generated data at each time step, but we omit this in our formulation for analytic simplicity. }Throughout our analysis we assume that all features are deterministic. We represent dataset $\tilde{D}_k$ with $\tilde{n}_k \times d$ matrix 
$\tilde{X}_k = [\tilde{x}_{k1}, \dots, \tilde{x}_{k \tilde{n}_k}]^\top $ and $\tilde{n}_k \times 1$ vector $\tilde{y}_k = [\tilde{y}_{k1}, \dots, y_{k\tilde{n}_k}]^\top$, and use the same convention for the public data $(X_* , y_*)$ and the generated data $(X_{tk}, y_{tk})$. Data across different entities are then combined  into ``lifted'' representations, which are denoted using boldface:
\begin{align}
\tilde{\bX} = \begin{bsmallmatrix}\tilde{X}_1 \\ & \ddots \\ && \tilde{X}_K \end{bsmallmatrix} , \quad \by_0  =  \begin{bsmallmatrix}\tilde{y}_1 \\  \vdots \\  \tilde{y}_K   \end{bsmallmatrix} , \quad \bX_t = \begin{bsmallmatrix}X_{t1} \\ & \ddots \\ && X_{tK} \end{bsmallmatrix} , \quad \by_t  =  \begin{bsmallmatrix}y_{tk} \\  \vdots \\  y_{tK} \end{bsmallmatrix} 
\end{align}
We note that information about which entity produced which sample is required for the analysis, but is not used during the training, where all data from the same generation are treated interchangeably.

\para{Bias-Variance Decomposition.} We derive exact formulas for the mean and variance of the estimators at each stage of the workflow. Given the features  $(\tilde{\bX}, X_*, \bX_t)$ and learning weights $(\alpha_t, \beta_t)$  define 
\begin{alignat}{3}
\tilde{\bS} &=\diag(\tilde{S}_1, \dots, \tilde{S}_k) \coloneqq \tilde{\bX}^\top \tilde{\bX}, \qquad S_* \coloneqq X_*^\top X_*  &&\bS_t = \diag(S_{t1}, \dots, S_{tk}) \coloneqq \bX_t^\top \bX_t\\
\bG_{t} & \coloneqq  \bar{\alpha}_t \beta_t \tilde{\bS}  + \bar{\alpha}_t \bar{\beta}_t  ( \Id_K \otimes S_* ) + \alpha_t  ( \Id_K \otimes \underline{\bS}_t )&\quad & \underline{\bS}_t \coloneqq \textstyle \frac{1}{K}  \sum_{k=1}^K S_{tk}
\\
 \bP_t &\coloneqq \bar{\alpha}_t \bG^+_t   \begin{bmatrix}  \beta_t  \tilde{\bS}   &  \bar{\beta}_t  (\one_K \otimes S_*)  \end{bmatrix} &\qquad & \bQ_t \coloneqq \alpha_t \bG^+_t \Pi \bS_t
\end{alignat}
where  
$\Pi \coloneqq \frac{1}{K} (\one_{K \times K} \otimes \Id_d)$ is an orthogonal projection matrix and $\alpha_0 \equiv 0$. 

\begin{callout}
\begin{theorem}\label{thm:MtCt}
Conditional on the initial data  $D_0 \coloneqq (\tilde{D}_1, \dots, \tilde{D}_K, D_*)$, 
the estimates $\hat{\btheta}_t = \gvec( \hat{\theta}_{t1}, \dots, \hat{\theta}_{tK})$ are Gaussian with mean and variance 
\begin{align}
  \ex{ \hat{\btheta}_t  \mid D_0 } = \bM_t
  \begin{bmatrix} \tilde{\bX}^+ \tilde{\by}  \\   X_*^+ y_*  \end{bmatrix} , \qquad 
\cov( \hat{\btheta}_t  \mid  D_0 )  = \bC_t 
\end{align}
where the matrices $\bM_t$ and $\bC_t$ are defined recursively  with $\bM_0 = \bP_0$ and $\bC_0 = \bm{0}_{Kd \times Kd}$ and 
\begin{alignat}{3}
    \bM_t& = \bP_t + \bQ_t \bM_{t-1}, & \qquad   \bC_t &= \bQ_t (\sigma^2 \bS_t^+ +\bC_{t-1} )\bQ_{t}, \qquad t \ge 1.
\end{alignat}  
\end{theorem}
\end{callout}

Theorem~\ref{thm:MtCt} shows that the conditional mean of each estimate $\hat{\theta}_{tk}$  is a linear combination of the individual ordinary least squares (OLS) estimates $\tilde{X}_1^+ \tilde{y}_1,\dots,  \tilde{X}_K^+ \tilde{y}_K$ and $X_*^+y_*$ for the private data and public data, respectively. For each generation $t$,  similarity across entities can be assessed by comparing the rows of the $K \times (K+1)$ block partitioning of   $\bM_t$.  At initialization, the off-diagonal blocks for the private data are zeroed out, but in later stages, these blocks become nonzero thereby allowing private data to be shared across  entities. Homogenization (i.e., shrinkage towards a global consensus) occurs when row blocks are identical, and thus each entity has the same mean.

 For our next result we mimic the experimental setup in Section~\ref{sec:results} and assume that the initial data are generated from a Gaussian model with a common ground truth parameter and the heterogeneity across datasets arises from the differences in the features, i.e., the matrices $\tilde{S}_1, \dots, \tilde{S}_K$. 

\begin{callout}
\begin{theorem}\label{thm:bias_mse}
Suppose that the initial data are generated independently according to the model $
y \mid x \sim \normal(x^\top \theta, \sigma^2 \Id_d)$ 
where $\theta \in \bbR^d$ is a fixed parameter. 
If $\bG_1, \dots, \bG_t$ are full rank then
\begin{align}
\ex{ \hat{\btheta}_t  } = \left( \Id -  \bQ_t \cdots \bQ_1(\Id -  \bG_0 \bG_0^+)  \right) (\one_K \otimes \theta)
,\quad \cov( \hat{\btheta}_t) = \bM_t \begin{bmatrix}  \tilde{\bS}^+ & 0 \\ 0 & S_*^+ \end{bmatrix} \bM_t^\top + \bC_t
\end{align}

\end{theorem} 
\end{callout}

To help interpret this result, observe that if $\bG_0$ is full rank, then each initial estimate is unbiased, and unbiasedness persists throughout every stage of training. Conversely, if $\bG_0$ is rank deficient, then at least one (and possibly all) of the initial estimates is biased. Remarkably, Theorem~\ref{thm:bias_mse} shows that it may still be possible for all entities have vanishing bias, provided that $\bQ_t \cdots \bQ_s$ converges to zero. Specific conditions under which this occurs are considered in the next section. 

\para{Asymptotic Variance.} 
To provide a finer analysis of the training dynamics we now suppose that the weights and features satisfy $\alpha_t = \alpha$,  $\beta_t = \beta$, and   $\bS_t = \bS$ for $t \ge 1$. Setting $\bP = \bP_1$ and $\bQ =\bQ_1$, the matrices $\bM_t$ and $\bC_t$ defined in Theorem~\ref{thm:MtCt} can be expressed explicitly as 
\begin{align}
\bM_t  = \bQ^t \bP_0 +  \Big(  \sum_{s=0}^{t-1} \bQ^{s} \Big)  \bP , \qquad & \bC_t  = \sigma^2 \sum_{s=1}^{t} \bQ^{s}  \bS^+ \big(  \bQ^{s} \big)^\top \label{eq:MtCt_alt} 
\end{align} 
Classical results in matrix analysis \citep{higham2002accuracy}
imply that if the spectral radius of $\bQ$ is strictly less than one,  then these these matrices converge to well-defined limits $\bM$ and $\bC$ satisfying 
\begin{align}
\bM   \coloneqq (\Id - \bQ)^{-1} \bP, \qquad
\gvec( \bC) \coloneqq \sigma^2  ( \Id - \bQ \otimes \bQ)^{-1} \gvec( \bQ \bS^+ \bQ) \label{eq:MC}
\end{align}

The following result provides a sufficient condition for convergence in terms of the triple $(\tilde{\bS}, S_*, \bS)$.  In particular, if $\bS$ is proportional to $\tilde{\bS}$ then the condition is satisfied for all $0 \le \alpha < 1$ and $0 \le \beta \le 1$. Note that the boundary case $\alpha = 1$ corresponds to the recursive training setting of \citet{shumailov2024ai} where the variance increases linearly across generations, and thus convergence does not occur.

\begin{lemma}\label{lem:convergence_cond} Suppose that $\bS \propto   \lambda \tilde{\bS} + (1- \lambda) \textcolor{black}{(\Id_K \otimes  S_* )}$ for some $0 <  \lambda \le 1$. Then,  the spectral radius of  $\bQ$ is strictly less than one for all $0 \le \alpha < 1$ and $0< \beta \le \lambda$. 
\end{lemma}

We summarize our findings with the following characterization of the asymptotic variance: 
\begin{callout}
\begin{theorem}\label{thm:asymp_var} 
 Consider the setting of Theorem~\ref{thm:bias_mse} and suppose that $\alpha_t = \alpha$,  $\beta_t = \beta$, and   $\bS_t = \bS$ for $t \ge 1$. If $\bG =\bG_1$ has full rank and $\bQ$ has spectral radius strictly less than one, then
 \begin{align}
\ex{ \btheta_t}\xrightarrow{t \to \infty}   \one_K \otimes \theta, \qquad   \cov( \hat{\btheta}_t) \xrightarrow{t \to \infty}  \sigma^2 \bM \begin{bmatrix}  \tilde{\bS}^+ & 0 \\ 0 & S_*^+ \end{bmatrix} \bM^\top + \bC
\end{align}
where $\bM$ and $\bC$ are given by \eqref{eq:MC}. 
\end{theorem}
\end{callout}

\para{MSE and relative efficiency.} The expression for the mean and variance in Theorems~\ref{thm:bias_mse} and  \ref{thm:asymp_var} provide explicit formulas for the mean squared error (MSE) $\ex{ \| \hat{\theta}_{tk} - \theta\|^2 }$ of entity $k$  at each generation $t$  and the mean squared prediction error (MSPE) $\ex{ \| \tilde{X}_{m} (\hat{\theta}_{tk} - \theta)\|^2 }$ for entity $m$'s private feature matrix. 

\begin{figure}[t]
    \vspace{-0.2cm}
    \centering
    \includegraphics[width=0.28\linewidth]{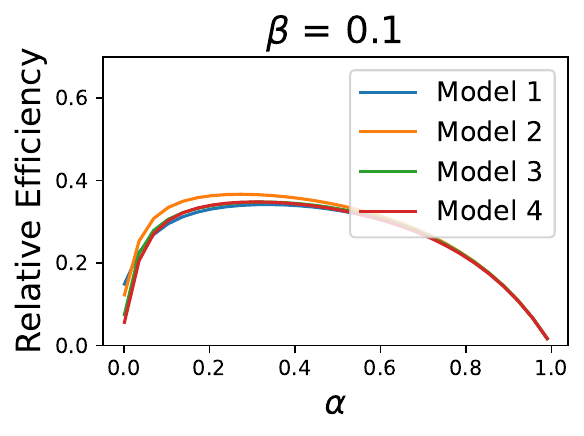}\hfill
    \includegraphics[width=0.28\linewidth]{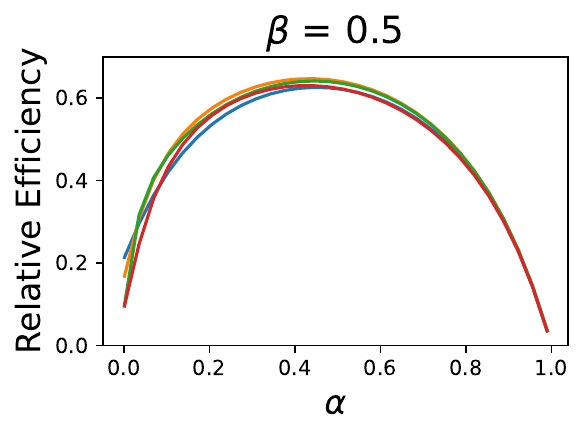}\hfill
    \includegraphics[width=0.28
    \linewidth]{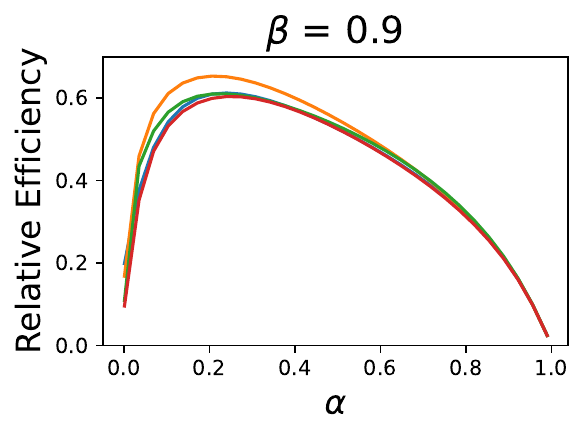}
    \vspace{-0.3cm}
    \caption{\small {\bf Predicted relative efficiency across $\alpha, \beta$ values for a $K=4$ model system (dimension $15$ and rank $5$).} 
     \em Curves show ratio of MSE of the minimum variance unbiased estimator to asymptotic MSE for a given $\alpha, \beta$ setting obtained from Theorem~\ref{thm:asymp_var} ($1$ is optimal). Setting $\alpha=\beta=0.5$ produces best results across models. 
}
    \label{fig:relative_efficiency}
    \vspace{-0.4cm}
\end{figure}

We can use this to compute the optimal $\alpha$, $\beta$ values for the interactive training setting. Figure~\ref{fig:relative_efficiency} compares the asymptotic MSE for a training workflow with given $\alpha$, $\beta$ (from Theorem~\ref{thm:asymp_var}) with the MSE of an idealized setting where each entity has access to the entire collection of real data (both private and public). Each curve represents the relative efficiency, i.e., the ratio of optimal MSE to entity-specific workflow MSE, with values close to one indicating near optimality. 
These results 
demonstrate that a setting with $\beta=0.5$ and $\alpha=0.5$ achieves the best global performance for all models. When $\beta$ is much larger (0.9), relatively small $\alpha$ values also improve model performance. 

\vspace{-0.3cm}
\section{Experimental Evaluation}
\label{sec:results}
\vspace{-.3cm}


To understand how our theoretical predictions bear out in practice,  we run experiments on text-generation models. In each, we train $K$ interacting models (per our framework in Figure~\ref{Fig: Interaction setup}) for several generations and evaluate how their performance changes on their own and other models' tasks. Here, we present the setup and results for a $K=2$ model system. Additional results are in the Appendix. 

\para{Experiment setup.} Training large language models from scratch is computationally infeasible for us, so we simulate the initial setup of language models trained on dataset  $(D_*,\tilde{D}_k)$ at time $t=0$ by fine-tuning $K=2$ instances of a given pre-trained model architecture on carefully chosen $D_*, \tilde{D}_k$. We experiment with two language model architectures\textemdash OPT-350m~\citep{zhang2022opt} and Llama 3.2 1B~\citep{dubey2024llama}\textemdash to assess how interactive training scales. Public sources~\citep{touvron2023llama, zhang2022opt} state that both models were pretrained on BookCorpus \citep{samuelyangbookcorpus}, CC-Stories \citep{andersonbcdefg}, the English portion of CommonCrawl, and public Reddit data. We approximate $D_*$ with BookCorpus due to practical constraints. Each model is given its own initial task-specific dataset $\tilde{D}_k$ \textemdash SciQ~\citep{SciQ} (science questions) for the $k=0$ model; and OpenAI's GSM8K~\citep{cobbe2021gsm8k} (grade school level math problems) for the $k=1$ model.

Interactive training proceeds as outlined in Figure~\ref{Fig: Interaction setup}, with fixed $\alpha$, $\beta$ values for each experiment. After training a new model generation $\hat{\theta}_{tk}$, we use $\hat{\theta}_{tk}$ to produce synthetic data $D_{t+1,k}$ that becomes part of the next generation's training data (if $\alpha>0$). We produce $D_{t+1, k}$ by randomly sampling prompts from $\tilde{D}_k$ for each of the $K$ models and prompting $\hat{\theta}_{tk}$ to complete the text.

\para{Training and evaluation.} We run experiments on $K=2$ model systems with $\alpha \in \{0, 0.5, 1\}$ and $\beta \in  \{0, 0.5, 1\}$, each for $T = 15$ generations of training. At each training generation,  models are fine-tuned on datasets of fixed size $n = 12,500$ drawn i.i.d.\ from the datasets $\tilde{D}_k$, $D_*$, $D_t$ with weights $\bar{\alpha} \beta$, $\bar{\alpha} \bar{\beta}$, and $\alpha/K$, respectively. This effectively mimics the {\em accumulate and subsample} setup of~\citet{kazdan2024collapse} with the additional wrinkle of data-mediated model interactions.

We train each model for $100$ steps per generation on a single NVIDIA H200 GPU using mixed-precision, the AdamW optimizer with a learning rate of 8$e^{-6}$, warmup ratio of 0.025, and gradient accumulation over 2 steps. 
After training each generation, we record evaluation loss (token-wise average cross-entropy loss) by feeding each model prompts from each test set of private data $\tilde{D}_k$ and evaluating semantic ``distance'' between predicted and correct answer. We also compute embedded representations of models' completions of the first $200$ elements of each of $\tilde{D}_k$ and $D_*$, so we can measure how various $\alpha$, $\beta$ affect models' representational spaces. Embeddings are computed via the \texttt{SentenceTransformers} python library. If models produce outputs with similar embeddings (as measured by cosine similarity), their feature spaces are more aligned. These two metrics allow us to evaluate how data-mediated interactions between models affect (1) models' performance on their own and other models' tasks and (2) model homogeneity as measured by embedding closeness.

\begin{table*}[h]
\vspace{-0.2cm}
\caption{\small {\bf Change in loss behavior for $K=2$ interacting models at $\beta=0.5$.} \em We show results as $\texttt{initial} \rightarrow \texttt{final}$ prediction loss values for models on their own and the other models' tasks, at $\texttt{initial} = T=0$  and $\texttt{final} = T =15$ generations. For clarity, we colorize loss \increase{increase}, \decrease{decrease}, and \neutral{constancy} ($\Delta \le 0.1$).}
\label{tab:k2_results}
\vspace{-0.3cm}
\begin{subtable}{0.5\linewidth}
    \centering
    \caption{\em OPT models}
\resizebox{0.95\textwidth}{!}{%
\begin{tabular}{cccc}
\toprule
& $\alpha = 0$    & $\alpha = 0.5$    & $\alpha = 1.0$                          \\ \midrule
\begin{tabular}[c]{@{}c@{}}Model 1\\ on Task 1\end{tabular} & \neutral{$3.3 \rightarrow 3.3$} & \neutral{$3.3 \rightarrow 3.3$} & \increase{$3.3 \rightarrow 3.5$} \\  \midrule
\begin{tabular}[c]{@{}c@{}}Model 2\\ on Task 2\end{tabular} & \neutral{$1.8 \rightarrow 1.7$} & \neutral{$1.8 \rightarrow 1.7$} & \increase{$1.8 \rightarrow 2.2$} \\ \midrule \midrule 
\begin{tabular}[c]{@{}c@{}}Model 1 \\ on Task 2\end{tabular} & \increase{$ 3.1 \rightarrow 3.5$} & \decrease{$3.1 \rightarrow 1.8$} & \decrease{$ 3.1 \rightarrow 2.2$}  \\ \midrule
\begin{tabular}[c]{@{}c@{}}Model 2 \\ on Task 1\end{tabular} & \neutral{$ 5.1 \rightarrow 5.1 $} & \decrease{$5.1 \rightarrow 3.5$} & \decrease{$ 5.1 \rightarrow 3.5 $} \\ \bottomrule
\end{tabular}%
}



\end{subtable}
\begin{subtable}{0.5\linewidth}
    \centering
    \caption{\em Llama 3.1 models}
\resizebox{0.95\textwidth}{!}{%
\begin{tabular}{cccc}
\toprule
& $\alpha = 0$        & $\alpha = 0.5$       & $\alpha =1.0$                          \\ \midrule
\begin{tabular}[c]{@{}c@{}}Model 1\\ on Task 1\end{tabular} & \neutral{$2.8 \rightarrow 2.9$} & \neutral{$2.8 \rightarrow 2.9$} & \increase{$2.8 \rightarrow 3.0$} \\  \midrule
\begin{tabular}[c]{@{}c@{}}Model 2\\ on Task 2\end{tabular} & \neutral{$1.2 \rightarrow 1.2$} & \neutral{$1.2 \rightarrow 1.3$} & \increase{$1.2 \rightarrow 1.9$} \\ \midrule \midrule
\begin{tabular}[c]{@{}c@{}}Model 1 \\ on Task 2\end{tabular} & \increase{$ 2.0 \rightarrow 2.5$}  & \decrease{$ 2.0 \rightarrow 1.4$} & \decrease{$ 2.0 \rightarrow 1.8$} \\  \midrule
\begin{tabular}[c]{@{}c@{}}Model 2 \\ on Task 1\end{tabular} & \increase{$ 3.7 \rightarrow 4.2 $} & \decrease{$ 3.7 \rightarrow 3.0$} & \decrease{$ 3.7 \rightarrow 3.0$} \\ \bottomrule
\end{tabular}%
}%


\end{subtable}
\vspace{-0.2cm}
\end{table*}



\begin{figure*}[h]
    \centering \includegraphics[width=0.95\textwidth]{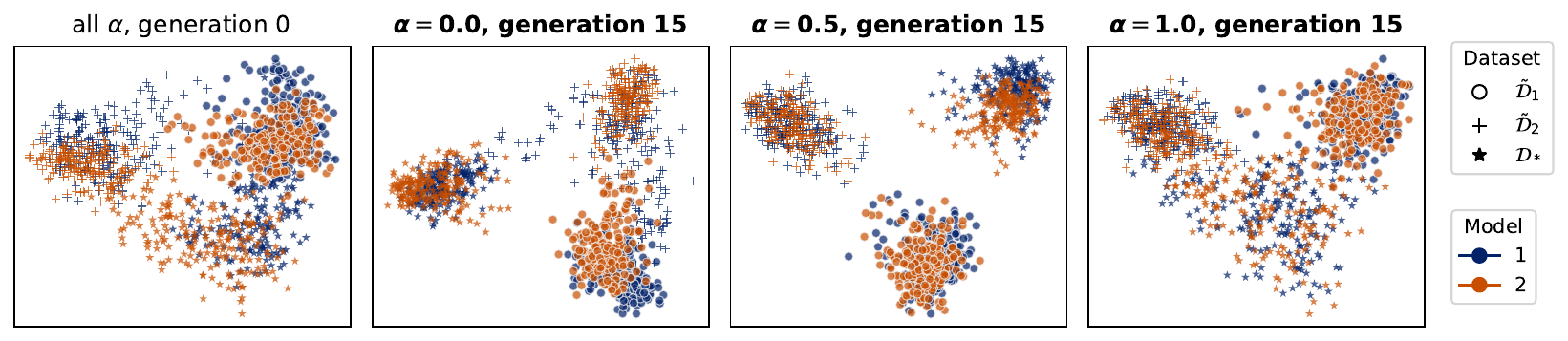}
    \vspace{-0.3cm}
    \caption{\small \em {\bf PCA of embeddings of outputs produced by models $\theta_{t1}$ and $\theta_{t2}$ on datasets $\tilde{\mathcal{D}}_1$, $\tilde{\mathcal{D}}_2$, and $\tilde{\mathcal{D}}_*$. Results for $K=2$ Llama models trained with $\beta=0.5$ and varying $\alpha$.} Leftmost plot shows embeddings at generation $t=0$, which are identical for all $\alpha$; while right plots show embeddings at $t=15$ with different $\alpha$. }
    \label{fig:pca_beta05}
    \vspace{-0.1cm}
\end{figure*}

\begin{table}[t]
\centering
\vspace{-0.1cm}
\caption{\small {\bf Cosine similarity of embedded output representations for $K=2$ Llama models at $\beta=0.5$.} \em We show $\texttt{initial(t=0)} \rightarrow \texttt{final(t=15)}$ for $\tilde{\mathcal{D}}_1$ $\tilde{\mathcal{D}}_2$, and $\mathcal{D}_*$. We colorize \increase{increase} and \decrease{decrease}.} 
\label{tab:sim_scores_beta05}
\vspace{-0.1cm}
\resizebox{0.5\textwidth}{!}{%
\begin{tabular}{cccc}
\toprule
       & $\alpha = 0$                           & $\alpha = 0.5$                         & $\alpha = 1.0$                          \\ \midrule
$\tilde{\mathcal{D}}_1$ & \decrease{$0.73 \rightarrow 0.70$} & \increase{$0.73 \rightarrow 0.88$} & \increase{$0.73 \rightarrow 0.86$} \\
$\tilde{\mathcal{D}}_2$  & \decrease{$0.88 \rightarrow 0.79$} & \increase{$0.88 \rightarrow 0.93$} & \increase{$0.88 \rightarrow 0.94$} \\  
${\mathcal{D}_*}$&  \increase{$0.50 \rightarrow 0.59$} & \increase{$0.50 \rightarrow 0.64$} & \increase{$0.50 \rightarrow 0.58$} \\ \bottomrule
\end{tabular}%
}
\vspace{-0.4cm}

\end{table}
%

\para{Results.} As predicted in Figure~\ref{fig:relative_efficiency}, an $\alpha=0.5$, $\beta=0.5$ setting produces optimal results in terms of model performance across tasks. Table~\ref{tab:k2_results} reports the change in models' loss values between the first and last ($t=15$) training generation for the $\beta=0.5$ setting. When $\alpha=0$, the same (human-generated) data is used for each training update, resulting in stable or slightly worse performance on different tasks, as models are forced into a local minima. When $\alpha=1.0$, models are only trained on generated outputs. They degrade on their original task due to the lack of real data but improve slightly on the other model's task. However, at $\alpha=0.5$, models perform well on their own tasks and improve on the other model's task.  Results for other $\beta$ and $K=3$ are in Appendix and echo findings here. 


While some amount of mixing improves model performance on previously-unseen tasks, homogenization occurs for $D_*$ at all $\alpha$ and for $\tilde{D}_k$ when $\alpha > 0$\textemdash everywhere it can. Figure~\ref{fig:pca_beta05} shows the benefit of interactive training when $\alpha, \beta$ are reasonable, visualizing PCA-reduced embeddings from $\tilde{D}_1$, $\tilde{D}_2$, and $D_*$ for $\beta=0.5$ and varying $\alpha$ as training progresses For $\alpha = 0.5$, the PCA shows clearly separated task clusters, indicating that both models are better tailored to the individual tasks.
In contrast, the lack of well-defined clusters when $\alpha \ne 0.5$ suggests reduced sensitivity to task type, e.g. generic answers. 
Yet, in Table~\ref{tab:sim_scores_beta05}, we see that when $\alpha=0$, models homogenize slightly on the shared task $D_*$ but diverge on model-specific tasks. This makes sense since models the models do not train on each other’s tasks. Once $\alpha > 0$, homogenization increases for all datasets/tasks.

\vspace{-0.3cm}
\section{Discussion}
\label{sec:discussion}
\vspace{-0.2cm}

\para{Limitations.} Our work has several limitations. We discuss possible objections to our claims about the increasing presence of generated outputs in training datasets (e.g.~\citep{drayson2025machine}) in Appendix~\ref{appx:counterarguments}. Also, our theoretical framework assumes that new data in model updates is purely synthetic. In reality, if internet scrapes are used to create model update datasets, they will contain both synthetic and real data. Finally, we assume that model trainers use new scraped data for each model update but only reuse data from initial training. This assumption may limit the range of outcomes.

\para{Broader Impacts.} If data-mediated interactions homogenize generative models, causing them to coalesce on certain viewpoints, this could lead to pervasive bias in AI-generated content.~\citet{peterson2025ai} discusses this possibility, while ~\citet{wenger2025we} showed homogeneity across creative outputs from many LLMs, suggesting these homogenization effects may already be felt by models. Much future study is needed to evaluate the extent to which data-mediated interactions fuel homogeneity (as opposed to other causes) and develop mitigations.

\para{Conclusions and Future Work.} We provide a first look at possible outcomes of genAI models trained on each others' data and find mixed effects. Training on other models' data exposes models to concepts possibly missed in their own training data, but can homogenize model behaviors. Future work could consider additional nuances of interactions between models, explore how these interactions evolve in other modalities like image generation, and investigate whether fixed points (e.g. like the universal $\pi^2/6$ pathway of~\citep{dey2024universality}) exist under this paradigm.

\newpage
\bibliographystyle{iclr2026_conference}
\bibliography{refs.bib}

\begin{thebibliography}{71}
\providecommand{\natexlab}[1]{#1}
\providecommand{\url}[1]{\texttt{#1}}
\expandafter\ifx\csname urlstyle\endcsname\relax
  \providecommand{\doi}[1]{doi: #1}\else
  \providecommand{\doi}{doi: \begingroup \urlstyle{rm}\Url}\fi

\bibitem[gen(2022)]{genai_creative}
{How Generative AI is Changing Creative Work}.
\newblock \emph{Harvard Business Review}, 2022.
\newblock \url{https://hbr.org/2022/11/how-generative-ai-is-changing-creative-work}.

\bibitem[app(2024)]{apple_ad}
Apple intelligence | writing tools | iphone 16, 2024.
\newblock \url{https://www.youtube.com/watch?v=3m0MoYKwVTM}.

\bibitem[coh(2024)]{cohere}
Command r and command r plus model card, 2024.
\newblock \url{https://docs.cohere.com/docs/responsible-use}.

\bibitem[Abdin et~al.(2023)Abdin, Aneja, Bubec, César, Mendes, Chen, et~al.]{phi2}
Marah Abdin, Jyoti Aneja, Sebastien Bubec, Caio César, Teodoro Mendes, Weizhu Chen, et~al.
\newblock Phi-2: The surprising power of small language models, 2023.
\newblock \url{https://www.microsoft.com/en-us/research/blog/phi-2-the-surprising-power-of-small-language-models/}.

\bibitem[Abdin et~al.(2024)Abdin, Aneja, Awadalla, Awadallah, Awan, Bach, Bahree, Bakhtiari, Bao, Behl, et~al.]{abdin2024phi}
Marah Abdin, Jyoti Aneja, Hany Awadalla, Ahmed Awadallah, Ammar~Ahmad Awan, Nguyen Bach, Amit Bahree, Arash Bakhtiari, Jianmin Bao, Harkirat Behl, et~al.
\newblock Phi-3 technical report: A highly capable language model locally on your phone.
\newblock \emph{arXiv preprint arXiv:2404.14219}, 2024.

\bibitem[Achiam et~al.(2023)Achiam, Adler, Agarwal, Ahmad, Akkaya, Aleman, Almeida, Altenschmidt, Altman, Anadkat, et~al.]{achiam2023gpt}
Josh Achiam, Steven Adler, Sandhini Agarwal, Lama Ahmad, Ilge Akkaya, Florencia~Leoni Aleman, Diogo Almeida, Janko Altenschmidt, Sam Altman, Shyamal Anadkat, et~al.
\newblock Gpt-4 technical report.
\newblock \emph{arXiv preprint arXiv:2303.08774}, 2023.

\bibitem[AI(2024{\natexlab{a}})]{llama_users}
Meta AI.
\newblock {The Future of AI: Built with LLama}, 2024{\natexlab{a}}.
\newblock \url{https://ai.meta.com/blog/future-of-ai-built-with-llama/}.

\bibitem[AI(2025)]{llama4}
Meta AI.
\newblock Llama 4 model card, 2025.
\newblock \url{https://github.com/meta-llama/llama-models/blob/main/models/llama4/MODEL_CARD.md}.

\bibitem[AI(2024{\natexlab{b}})]{oai_watermark}
Open AI.
\newblock {Understanding the source of what we see and hear online}, 2024{\natexlab{b}}.
\newblock \url{https://openai.com/index/understanding-the-source-of-what-we-see-and-hear-online/}.

\bibitem[Alemohammad et~al.(2024)Alemohammad, Casco-Rodriguez, Luzi, Humayun, Babaei, LeJeune, Siahkoohi, and Baraniuk]{alemohammad2023self}
Sina Alemohammad, Josue Casco-Rodriguez, Lorenzo Luzi, Ahmed~Imtiaz Humayun, Hossein Babaei, Daniel LeJeune, Ali Siahkoohi, and Richard~G Baraniuk.
\newblock Self-consuming generative models go mad.
\newblock \emph{Proc. of ICLR}, 2024.

\bibitem[Anderson(2022)]{andersonbcdefg}
Benjamin Anderson.
\newblock Cc-stories, 2022.
\newblock \url{https://huggingface.co/datasets/andersonbcdefg/cc-stories-parquet/commits/main}.

\bibitem[Anthropic()]{claude_rlhf}
Anthropic.
\newblock Dataset card for hh-rlhf.
\newblock \url{https://huggingface.co/datasets/Anthropic/hh-rlhf}.

\bibitem[Anthropic(2023)]{claude2}
Anthropic.
\newblock Model card and evaluations for claude models, 2023.
\newblock \url{https://www-cdn.anthropic.com/bd2a28d2535bfb0494cc8e2a3bf135d2e7523226/Model-Card-Claude-2.pdf}.

\bibitem[Barzilai \& Shamir(2025)Barzilai and Shamir]{barzilai2025models}
Daniel Barzilai and Ohad Shamir.
\newblock When models don't collapse: On the consistency of iterative mle.
\newblock \emph{arXiv preprint arXiv:2505.19046}, 2025.

\bibitem[Brinner et~al.(2025)Brinner, Mustafa, and Zarrie{\ss}]{brinner2025enhancing}
Marc Brinner, Tarek~Al Mustafa, and Sina Zarrie{\ss}.
\newblock Enhancing domain-specific encoder models with llm-generated data: How to leverage ontologies, and how to do without them.
\newblock \emph{arXiv preprint arXiv:2503.22006}, 2025.

\bibitem[Brown et~al.(2020)Brown, Mann, Ryder, Subbiah, Kaplan, et~al.]{brown2020language}
Tom~B Brown, Benjamin Mann, Nick Ryder, Melanie Subbiah, Jared Kaplan, et~al.
\newblock Language models are few-shot learners.
\newblock \emph{Proc. of NeurIPS}, 2020.

\bibitem[Chowdhery et~al.(2023)Chowdhery, Narang, Devlin, Bosma, Mishra, Roberts, Barham, Chung, Sutton, Gehrmann, et~al.]{chowdhery2023palm}
Aakanksha Chowdhery, Sharan Narang, Jacob Devlin, Maarten Bosma, Gaurav Mishra, Adam Roberts, Paul Barham, Hyung~Won Chung, Charles Sutton, Sebastian Gehrmann, et~al.
\newblock Palm: Scaling language modeling with pathways.
\newblock \emph{Journal of Machine Learning Research}, 2023.

\bibitem[Clegg(2024)]{meta_watermark}
Nick Clegg.
\newblock {Labeling AI-Generated Images on Facebook, Instagram and Threads}.
\newblock \emph{{Meta AI Blog}}, 2024.
\newblock \url{https://about.fb.com/news/2024/02/labeling-ai-generated-images-on-facebook-instagram-and-threads/}.

\bibitem[Cobbe et~al.(2021)Cobbe, Kosaraju, Bavarian, Chen, Jun, Kaiser, Plappert, Tworek, Hilton, Nakano, Hesse, and Schulman]{cobbe2021gsm8k}
Karl Cobbe, Vineet Kosaraju, Mohammad Bavarian, Mark Chen, Heewoo Jun, Lukasz Kaiser, Matthias Plappert, Jerry Tworek, Jacob Hilton, Reiichiro Nakano, Christopher Hesse, and John Schulman.
\newblock Training verifiers to solve math word problems.
\newblock \emph{arXiv preprint arXiv:2110.14168}, 2021.

\bibitem[Crawl(2025)]{commoncrawl}
Common Crawl.
\newblock {Common Crawl - Open Repository of Web Crawl Data.}, 2025.
\newblock \url{https://commoncrawl.org/}.

\bibitem[Dathathri et~al.(2024)Dathathri, See, Ghaisas, Huang, McAdam, Welbl, Bachani, Kaskasoli, Stanforth, Matejovicova, et~al.]{dathathri2024scalable}
Sumanth Dathathri, Abigail See, Sumedh Ghaisas, Po-Sen Huang, Rob McAdam, Johannes Welbl, Vandana Bachani, Alex Kaskasoli, Robert Stanforth, Tatiana Matejovicova, et~al.
\newblock Scalable watermarking for identifying large language model outputs.
\newblock \emph{Nature}, 634\penalty0 (8035), 2024.

\bibitem[Dey \& Donoho(2024)Dey and Donoho]{dey2024universality}
Apratim Dey and David Donoho.
\newblock Universality of the $\pi^{2}/6$ pathway in avoiding model collapse.
\newblock \emph{arXiv preprint arXiv:2410.22812}, 2024.

\bibitem[Dohmatob et~al.(2025)Dohmatob, Feng, and Kempe]{dohmatob2025model}
Elvis Dohmatob, Yunzhen Feng, and Julia Kempe.
\newblock Model collapse demystified: The case of regression.
\newblock \emph{Proc. of NeurIPS}, 2025.

\bibitem[Drayson et~al.(2025)Drayson, Yilmaz, and Lampos]{drayson2025machine}
George Drayson, Emine Yilmaz, and Vasileios Lampos.
\newblock Machine-generated text detection prevents language model collapse.
\newblock \emph{arXiv preprint arXiv:2502.15654}, 2025.

\bibitem[Du et~al.(2022)Du, Huang, Dai, Tong, Lepikhin, et~al.]{du2022glam}
Nan Du, Yanping Huang, Andrew~M Dai, Simon Tong, Dmitry Lepikhin, et~al.
\newblock Glam: Efficient scaling of language models with mixture-of-experts.
\newblock In \emph{Proc. of ICML}, 2022.

\bibitem[Dubey et~al.(2024)Dubey, Jauhri, Pandey, Kadian, Al-Dahle, Letman, Mathur, Schelten, Yang, Fan, et~al.]{dubey2024llama}
Abhimanyu Dubey, Abhinav Jauhri, Abhinav Pandey, Abhishek Kadian, Ahmad Al-Dahle, Aiesha Letman, Akhil Mathur, Alan Schelten, Amy Yang, Angela Fan, et~al.
\newblock The llama 3 herd of models.
\newblock \emph{arXiv preprint arXiv:2407.21783}, 2024.

\bibitem[Ellis(2024)]{grammarly_ad}
Matt Ellis.
\newblock How to use ai to enhance your storytelling process, 2024.
\newblock \url{https://www.grammarly.com/blog/writing-with-ai/ai-story-writing/}.

\bibitem[Feng et~al.(2024)Feng, Dohmatob, Yang, Charton, and Kempe]{feng2024beyond}
Yunzhen Feng, Elvis Dohmatob, Pu~Yang, Francois Charton, and Julia Kempe.
\newblock Beyond model collapse: Scaling up with synthesized data requires reinforcement.
\newblock In \emph{ICML Workshop on Theoretical Foundations of Foundation Models}, 2024.

\bibitem[GAO(2024)]{health_ai_1}
US~GAO.
\newblock Science and tech spotlight - generative ai in health care, 2024.
\newblock GAO-24-107634, \url{https://www.gao.gov/products/gao-24-107634}.

\bibitem[Gerstgrasser et~al.(2024)Gerstgrasser, Schaeffer, Dey, Rafailov, et~al.]{gerstgrasser2024model}
Matthias Gerstgrasser, Rylan Schaeffer, Apratim Dey, Rafael Rafailov, et~al.
\newblock Is model collapse inevitable? breaking the curse of recursion by accumulating real and synthetic data.
\newblock \emph{Proc. of COLM}, 2024.

\bibitem[Handa et~al.(2025)Handa, Tamkin, McCain, Huang, Durmus, Heck, Mueller, Hong, Ritchie, Belonax, et~al.]{handa2025economic}
Kunal Handa, Alex Tamkin, Miles McCain, Saffron Huang, Esin Durmus, Sarah Heck, Jared Mueller, Jerry Hong, Stuart Ritchie, Tim Belonax, et~al.
\newblock {Which Economic Tasks are Performed with AI? Evidence from Millions of Claude Conversations}.
\newblock \emph{arXiv preprint arXiv:2503.04761}, 2025.

\bibitem[Harding(2024)]{dod_ai_use_1}
Emily Harding.
\newblock {2024 Priorities for the Intelligence Community}.
\newblock \emph{Center for Strategic and International Studies}, 2024.
\newblock \url{https://www.csis.org/analysis/2024-priorities-intelligence-community-0}.

\bibitem[Hataya et~al.(2023)Hataya, Bao, and Arai]{hataya2023will}
Ryuichiro Hataya, Han Bao, and Hiromi Arai.
\newblock Will large-scale generative models corrupt future datasets?
\newblock In \emph{Proc. of ICCV}, 2023.

\bibitem[Higham(2002)]{higham2002accuracy}
Nicholas~J Higham.
\newblock \emph{Accuracy and stability of numerical algorithms}.
\newblock SIAM, 2002.

\bibitem[Hoffmann et~al.(2022)Hoffmann, Borgeaud, Mensch, Buchatskaya, Cai, Rutherford, Casas, Hendricks, Welbl, Clark, et~al.]{hoffmann2022training}
Jordan Hoffmann, Sebastian Borgeaud, Arthur Mensch, Elena Buchatskaya, Trevor Cai, Eliza Rutherford, Diego de~Las Casas, Lisa~Anne Hendricks, Johannes Welbl, Aidan Clark, et~al.
\newblock Training compute-optimal large language models.
\newblock \emph{Proc. of NeurIPS}, 2022.

\bibitem[Jain \& Krishnamurthy(2025)Jain and Krishnamurthy]{jain2025interacting}
Adit Jain and Vikram Krishnamurthy.
\newblock Interacting large language model agents. bayesian social learning based interpretable models.
\newblock \emph{IEEE Access}, 2025.

\bibitem[Jiang et~al.(2023)Jiang, Sablayrolles, Mensch, Bamford, Chaplot, de~las Casas, Bressand, Lengyel, Lample, Saulnier, et~al.]{jiang2023mistral}
AQ~Jiang, A~Sablayrolles, A~Mensch, C~Bamford, DS~Chaplot, D~de~las Casas, F~Bressand, G~Lengyel, G~Lample, L~Saulnier, et~al.
\newblock Mistral 7b (2023).
\newblock \emph{arXiv preprint arXiv:2310.06825}, 2023.

\bibitem[Johannes~Welbl(2017)]{SciQ}
Matt~Gardner Johannes~Welbl, Nelson F.~Liu.
\newblock Crowdsourcing multiple choice science questions.
\newblock 2017.

\bibitem[Kaplan et~al.(2020)Kaplan, McCandlish, Henighan, Brown, Chess, Child, Gray, Radford, Wu, and Amodei]{kaplan2020scaling}
Jared Kaplan, Sam McCandlish, Tom Henighan, Tom~B Brown, Benjamin Chess, Rewon Child, Scott Gray, Alec Radford, Jeffrey Wu, and Dario Amodei.
\newblock Scaling laws for neural language models.
\newblock \emph{arXiv preprint arXiv:2001.08361}, 2020.

\bibitem[Kazdan et~al.(2025)Kazdan, Schaeffer, Dey, Gerstgrasser, Rafailov, Donoho, and Koyejo]{kazdan2024collapse}
Joshua Kazdan, Rylan Schaeffer, Apratim Dey, Matthias Gerstgrasser, Rafael Rafailov, David~L Donoho, and Sanmi Koyejo.
\newblock Collapse or thrive? perils and promises of synthetic data in a self-generating world.
\newblock \emph{Proc. of ICML}, 2025.

\bibitem[Kim et~al.(2022)Kim, Mishra, Jin, Panda, Kuehne, Karlinsky, Saligrama, Saenko, Oliva, and Feris]{kim2022transferable}
Yo-whan Kim, Samarth Mishra, SouYoung Jin, Rameswar Panda, Hilde Kuehne, Leonid Karlinsky, Venkatesh Saligrama, Kate Saenko, Aude Oliva, and Rogerio Feris.
\newblock How transferable are video representations based on synthetic data?
\newblock \emph{Proc. of NeurIPS}, 2022.

\bibitem[Li et~al.(2023)Li, Bubeck, Eldan, Del~Giorno, Gunasekar, and Lee]{li2023textbooks}
Yuanzhi Li, S{\'e}bastien Bubeck, Ronen Eldan, Allie Del~Giorno, Suriya Gunasekar, and Yin~Tat Lee.
\newblock Textbooks are all you need ii: phi-1.5 technical report.
\newblock \emph{arXiv preprint arXiv:2309.05463}, 2023.

\bibitem[Longpre et~al.(2024)Longpre, Mahari, Obeng-Marnu, Brannon, South, Kabbara, and Pentland]{longpre2024data}
Shayne Longpre, Robert Mahari, Naana Obeng-Marnu, William Brannon, Tobin South, Jad Kabbara, and Sandy Pentland.
\newblock Data authenticity, consent, and provenance for ai are all broken: What will it take to fix them?
\newblock 2024.

\bibitem[Marchi et~al.(2024)Marchi, Soatto, Chaudhari, and Tabuada]{marchi2024heat}
Matteo Marchi, Stefano Soatto, Pratik Chaudhari, and Paulo Tabuada.
\newblock Heat death of generative models in closed-loop learning.
\newblock \emph{Proc. of IEEE 63rd Conference on Decision and Control (CDC)}, 2024.

\bibitem[Mart{\'\i}nez et~al.(2023)Mart{\'\i}nez, Watson, Reviriego, Hern{\'a}ndez, Juarez, and Sarkar]{martinez2023towards}
Gonzalo Mart{\'\i}nez, Lauren Watson, Pedro Reviriego, Jos{\'e}~Alberto Hern{\'a}ndez, Marc Juarez, and Rik Sarkar.
\newblock Towards understanding the interplay of generative artificial intelligence and the internet.
\newblock In \emph{International Workshop on Epistemic Uncertainty in Artificial Intelligence}. Springer, 2023.

\bibitem[Matatov et~al.(2024)Matatov, Qu{\'e}r{\'e}, Amir, and Naaman]{matatov2024examining}
Hana Matatov, Marianne Aubin~Le Qu{\'e}r{\'e}, Ofra Amir, and Mor Naaman.
\newblock Examining the prevalence and dynamics of ai-generated media in art subreddits.
\newblock \emph{arXiv preprint arXiv:2410.07302}, 2024.

\bibitem[Notion(2024)]{notion_ad}
Notion.
\newblock Use notion ai to write better, more efficient notes and docs, 2024.
\newblock \url{https://www.notion.com/help/guides/notion-ai-for-docs}.

\bibitem[Peterson(2025)]{peterson2025ai}
Andrew~J Peterson.
\newblock Ai and the problem of knowledge collapse.
\newblock \emph{AI \& SOCIETY}, pp.\  1--21, 2025.

\bibitem[Radford et~al.(2018)Radford, Narasimhan, Salimans, Sutskever, et~al.]{radford2018improving}
Alec Radford, Karthik Narasimhan, Tim Salimans, Ilya Sutskever, et~al.
\newblock Improving language understanding by generative pre-training.
\newblock 2018.

\bibitem[Radford et~al.(2019)Radford, Wu, Child, Luan, Amodei, Sutskever, et~al.]{radford2019language}
Alec Radford, Jeffrey Wu, Rewon Child, David Luan, Dario Amodei, Ilya Sutskever, et~al.
\newblock Language models are unsupervised multitask learners.
\newblock \emph{OpenAI blog}, 1\penalty0 (8):\penalty0 9, 2019.

\bibitem[Reddy(2024)]{reddy2024generative}
Sandeep Reddy.
\newblock Generative ai in healthcare: an implementation science informed translational path on application, integration and governance.
\newblock \emph{Implementation Science}, 19\penalty0 (1):\penalty0 27, 2024.

\bibitem[Reuters(2024)]{reuters_openai}
Reuters.
\newblock {OpenAI says ChatGPT's weekly users have grown to 200 million}, 2024.
\newblock \url{https://www.reuters.com/technology/artificial-intelligence/openai-says-chatgpts-weekly-users-have-grown-200-million-2024-08-29/}.

\bibitem[Sadasivan et~al.(2023)Sadasivan, Kumar, Balasubramanian, Wang, and Feizi]{sadasivan2023can}
Vinu~Sankar Sadasivan, Aounon Kumar, Sriram Balasubramanian, Wenxiao Wang, and Soheil Feizi.
\newblock Can ai-generated text be reliably detected?
\newblock \emph{arXiv preprint arXiv:2303.11156}, 2023.

\bibitem[Schaeffer et~al.(2025)Schaeffer, Kazdan, Arulandu, and Koyejo]{schaeffer2025position}
Rylan Schaeffer, Joshua Kazdan, Alvan~Caleb Arulandu, and Sanmi Koyejo.
\newblock Position: Model collapse does not mean what you think.
\newblock \emph{arXiv preprint arXiv:2503.03150}, 2025.

\bibitem[Shumailov et~al.(2024)Shumailov, Shumaylov, Zhao, Papernot, Anderson, and Gal]{shumailov2024ai}
Ilia Shumailov, Zakhar Shumaylov, Yiren Zhao, Nicolas Papernot, Ross Anderson, and Yarin Gal.
\newblock Ai models collapse when trained on recursively generated data.
\newblock \emph{Nature}, 631\penalty0 (8022):\penalty0 755--759, 2024.

\bibitem[Sun et~al.(2025)Sun, Zhang, Shen, Zhang, Liu, Backes, Zhang, and He]{sun2024we}
Zhen Sun, Zongmin Zhang, Xinyue Shen, Ziyi Zhang, Yule Liu, Michael Backes, Yang Zhang, and Xinlei He.
\newblock {Are We in the AI-Generated Text World Already? Quantifying and Monitoring AIGT on Social Media}.
\newblock \emph{Proc. of ACL}, 2025.

\bibitem[Team et~al.(2024{\natexlab{a}})Team, Georgiev, Lei, Burnell, Bai, Gulati, Tanzer, Vincent, Pan, Wang, et~al.]{team2024gemini}
Gemini Team, Petko Georgiev, Ving~Ian Lei, Ryan Burnell, Libin Bai, Anmol Gulati, Garrett Tanzer, Damien Vincent, Zhufeng Pan, Shibo Wang, et~al.
\newblock Gemini 1.5: Unlocking multimodal understanding across millions of tokens of context.
\newblock \emph{arXiv preprint arXiv:2403.05530}, 2024{\natexlab{a}}.

\bibitem[Team et~al.(2024{\natexlab{b}})Team, Lenz, Arazi, Bergman, Manevich, Peleg, Aviram, Almagor, Fridman, Padnos, et~al.]{team2024jamba}
Jamba Team, Barak Lenz, Alan Arazi, Amir Bergman, Avshalom Manevich, Barak Peleg, Ben Aviram, Chen Almagor, Clara Fridman, Dan Padnos, et~al.
\newblock Jamba-1.5: Hybrid transformer-mamba models at scale.
\newblock \emph{arXiv preprint arXiv:2408.12570}, 2024{\natexlab{b}}.

\bibitem[Thoppilan et~al.(2022)Thoppilan, De~Freitas, Hall, Shazeer, Kulshreshtha, Cheng, Jin, Bos, Baker, Du, et~al.]{thoppilan2022lamda}
Romal Thoppilan, Daniel De~Freitas, Jamie Hall, Noam Shazeer, Apoorv Kulshreshtha, Heng-Tze Cheng, Alicia Jin, Taylor Bos, Leslie Baker, Yu~Du, et~al.
\newblock Lamda: Language models for dialog applications.
\newblock \emph{arXiv preprint arXiv:2201.08239}, 2022.

\bibitem[Tian \& Shen(2025)Tian and Shen]{tian2025generative}
Xinyu Tian and Xiaotong Shen.
\newblock Generative distribution prediction: A unified approach to multimodal learning.
\newblock \emph{arXiv preprint arXiv:2502.07090}, 2025.

\bibitem[Touvron et~al.(2023{\natexlab{a}})Touvron, Lavril, Izacard, Martinet, Lachaux, Lacroix, Rozi{\`e}re, Goyal, Hambro, Azhar, et~al.]{touvron2023llama}
Hugo Touvron, Thibaut Lavril, Gautier Izacard, Xavier Martinet, Marie-Anne Lachaux, Timoth{\'e}e Lacroix, Baptiste Rozi{\`e}re, Naman Goyal, Eric Hambro, Faisal Azhar, et~al.
\newblock Llama: Open and efficient foundation language models.
\newblock \emph{arXiv preprint arXiv:2302.13971}, 2023{\natexlab{a}}.

\bibitem[Touvron et~al.(2023{\natexlab{b}})Touvron, Martin, Stone, Albert, Almahairi, Babaei, Bashlykov, Batra, Bhargava, Bhosale, et~al.]{touvron2023llama_b}
Hugo Touvron, Louis Martin, Kevin Stone, Peter Albert, Amjad Almahairi, Yasmine Babaei, Nikolay Bashlykov, Soumya Batra, Prajjwal Bhargava, Shruti Bhosale, et~al.
\newblock Llama 2: Open foundation and fine-tuned chat models.
\newblock \emph{arXiv preprint arXiv:2307.09288}, 2023{\natexlab{b}}.

\bibitem[Wang et~al.(2024)Wang, Wu, Zhang, Jain, Guan, and Koshiyama]{wang2024bias}
Ze~Wang, Zekun Wu, Jeremy Zhang, Navya Jain, Xin Guan, and Adriano Koshiyama.
\newblock Bias amplification: Language models as increasingly biased media.
\newblock \emph{arXiv preprint arXiv:2410.15234}, 2024.

\bibitem[Wenger \& Kenett(2025)Wenger and Kenett]{wenger2025we}
Emily Wenger and Yoed Kenett.
\newblock We're different, we're the same: Creative homogeneity across llms.
\newblock \emph{arXiv preprint arXiv:2501.19361}, 2025.
\newblock \url{https://arxiv.org/abs/2501.19361}.

\bibitem[xAI(2025)]{grok3}
xAI.
\newblock {Grok 3 Beta - The Age of Reasoning Agents}, 2025.
\newblock \url{https://x.ai/news/grok-3}.

\bibitem[Yang(2022)]{samuelyangbookcorpus}
Smuel Yang.
\newblock Bookcorpus dataset, 2022.
\newblock \url{https://huggingface.co/datasets/SamuelYang/bookcorpus}.

\bibitem[Zhang et~al.(2023)Zhang, Edelman, Francati, Venturi, Ateniese, and Barak]{zhang2023watermarks}
Hanlin Zhang, Benjamin~L Edelman, Danilo Francati, Daniele Venturi, Giuseppe Ateniese, and Boaz Barak.
\newblock Watermarks in the sand: Impossibility of strong watermarking for generative models.
\newblock \emph{arXiv preprint arXiv:2311.04378}, 2023.

\bibitem[Zhang et~al.(2024)Zhang, Qiao, Yang, and Wei]{zhang2024regurgitative}
Jinghui Zhang, Dandan Qiao, Mochen Yang, and Qiang Wei.
\newblock Regurgitative training: The value of real data in training large language models.
\newblock \emph{arXiv preprint arXiv:2407.12835}, 2024.

\bibitem[Zhang et~al.(2022)Zhang, Roller, Goyal, Artetxe, Chen, Chen, Dewan, Diab, Li, Lin, Mihaylov, Ott, Shleifer, Shuster, Simig, Koura, Sridhar, Wang, and Zettlemoyer]{zhang2022opt}
Susan Zhang, Stephen Roller, Naman Goyal, Mikel Artetxe, Moya Chen, Shuohui Chen, Christopher Dewan, Mona Diab, Xian Li, Xi~Victoria Lin, Todor Mihaylov, Myle Ott, Sam Shleifer, Kurt Shuster, Daniel Simig, Punit~Singh Koura, Anjali Sridhar, Tianlu Wang, and Luke Zettlemoyer.
\newblock Opt: Open pre-trained transformer language models, 2022.

\bibitem[Zhuang et~al.(2020)Zhuang, Qi, Duan, Xi, Zhu, Zhu, Xiong, and He]{zhuang2020comprehensive}
Fuzhen Zhuang, Zhiyuan Qi, Keyu Duan, Dongbo Xi, Yongchun Zhu, Hengshu Zhu, Hui Xiong, and Qing He.
\newblock A comprehensive survey on transfer learning.
\newblock \emph{Proceedings of the IEEE}, 109\penalty0 (1):\penalty0 43--76, 2020.

\bibitem[Zitron(2025)]{genai_revolution}
Edward Zitron.
\newblock {There is No AI Revolution}, 2025.
\newblock \url{https://www.wheresyoured.at/wheres-the-money/}.

\end{thebibliography}

\newpage
\appendix
\begin{center}
{\large \bf Appendix}
\end{center}

\section{Exact training data statements from LLM papers}

Table~\ref{tab:training_info} lists statements made about model training and fine-tuning data for large-scale generative AI models that do not explicitly list training data sources. 

\begin{table*}[h]
\centering
\resizebox{\textwidth}{!}{%
\begin{tabular}{ccc}
\toprule
{\bf Model} & {\bf Pre-training data}  & {\bf Fine-tuning data} \\ \midrule
Claude 2~\citep{claude2} & \begin{tabular}{c} Claude models are trained on a proprietary mix\\of publicly available information from the Internet,\\ datasets that we license from third party businesses,\\ 
and data that our users affirmatively share\\or that crowd workers provide. \end{tabular} & Publicly released on HuggingFace~\citep{claude_rlhf}. \\ \midrule
GPT4+~\citep{achiam2023gpt} & No information provided & No information provided \\  \midrule
Grok 3~\citep{grok3} & No information provided & No information provided \\ \midrule
 Jamba~\citep{team2024jamba}      &   \begin{tabular}{c} Our pre-training dataset is a mixture of\\ publicly available web documents,\\
code, books and scientific articles.\end{tabular} &   \begin{tabular}{c} When performing supervised fine-tuning,\\ we make heavy use of synthetic data.\end{tabular}    \\  \midrule
Llama 3~\citep{dubey2024llama} & \begin{tabular}{c} We create our dataset for language model pre-training\\ from a variety of data sources containing knowledge\\
until the end of 2023. Much of the data we utilize\\is obtained from the web  \end{tabular} & \begin{tabular}{c} We produce the aligned Llama 3 models by\\applying several rounds of post-training, or aligning\\the model
with human feedback.\end{tabular} \\ \midrule
Llama 4~\citep{llama4} & \begin{tabular}{c} A mix of publicly available, licensed data\\and information from Meta’s products and services.\\This includes publicly shared posts from Instagram and\\Facebook and people’s interactions with Meta AI.\end{tabular} & No information provided. \\ \midrule
Phi 3~\citep{abdin2024phi} & \begin{tabular}{c} Our training data of consists of\\heavily filtered
publicly available\\web data \ldots from various open internet sources,\\ as
well as synthetic LLM-generated data.\end{tabular}& \begin{tabular}{c}$[$Supervised fine tuning$]$ leverages highly\\curated high-quality data across\\diverse
domains, e.g., math, coding,\\reasoning, conversation, model identity, and safety.\end{tabular} \\ 
\bottomrule
\end{tabular}
}
\caption{{\bf Exact wording of training and fine-tuning data discussion from whitepapers in which data sources are not explicitly listed.}}
\label{tab:training_info}
\end{table*}

\section{Counterarguments}
\label{appx:counterarguments}

We argue that AI-generated content from a variety of sources will be increasingly prevalent online, resulting in future genAI models being regularly trained on each other's outputs. We believe this vision of future data-mediated interactions between models is reasonable, based on evidence from academic literature and corporate reports. However, others may disagree with our argument. Here, we present possible counterarguments to our view to catalyze future work and discussion. 

\para{Can't we use watermarks to filter AI-generated content from future internet-scraped datasets?} Several companies have publicly stated that they watermark AI-generated content~\citep{dathathri2024scalable, oai_watermark, meta_watermark}, making this argument plausible. Furthermore,~\citep{drayson2025machine} show that using watermark detection techniques can help avoid model collapse under certain circumstances. However, reliance on watermarking has two major issues. First, watermarks are difficult to reliably detect and/or easily removed from generated content~\citep{zhang2023watermarks,longpre2024data, sadasivan2023can}. Second, detection via watermark requires sharing of watermark detection information, which is essentially a game-theoretic problem that relies on other companies' willingness to cooperate. 
Both issues make watermarks an unreliable mechanism to rid datasets of AI-generated content. 

\para{What if there's less AI-generated online content than we think?} Some work suggests there may be less AI-generated content online than previously postulated~\citep{matatov2024examining}. However, other works consistently point to an uptick in AI-generated content online~\citep{sun2024we}. We believe the widespread adoption and use of generative AI models across industries~\citep{llama_users, handa2025economic, reuters_openai}, particularly for use in content creation~\citep{genai_creative}, provides strong evidence that AI-generated content will become a regular part of online life. Further empirical work is needed to vet both claims. 

\para{What if one model provider dominates online content?} Although numerous generative AI models are available online, some evidence suggests that one or two companies may dominate the AI landscape. A market research firm estimated that in January 2025, Open AI's ChatGPT had 340 million monthly active users, Microsoft Copilot had 11 million, Google Gemini had 80 million, and Anthropic's Claude had 2 million~\citep{genai_revolution}. If only one model/company dominates, our paradigm of training on other models' data will no longer be relevant and collapses back to the single model setting of prior work, e.g.~\citep{shumailov2024ai, kazdan2024collapse, dey2024universality}.  Currently, though, this market research suggests that there are several models used by millions of users, making our assumptions somewhat reasonable. Future research could analyze market trends in model use and adoption to determine realistic assumptions. 

\section{Additional Results}

\begin{figure}[ht]
    \centering
    \includegraphics[width=0.9\linewidth]{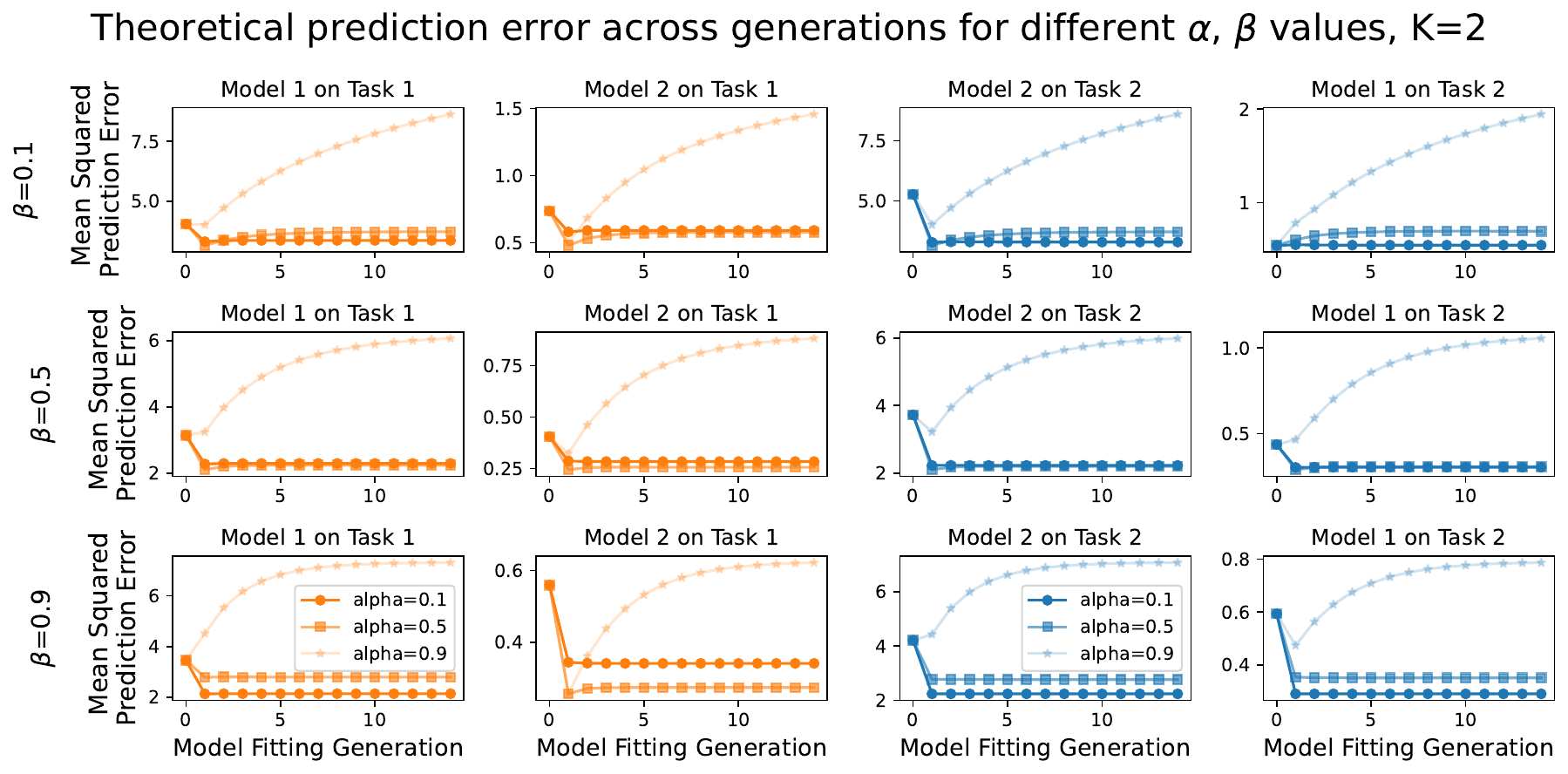}
    \vspace{-0.2cm}
    \caption{\small {\bf Predicted behavior over time for a $K=2$ model system with varying $\alpha$, $\beta$.} \em We use equations for MSE from theorem~\ref{thm:bias_mse} and run simulations with $K=2$, dimension 50, rank 15.}
    \label{fig:k=2_theory}
    \vspace{-0.1cm}
\end{figure}

\begin{figure}[ht]
    \centering
    \includegraphics[width=0.9\linewidth]{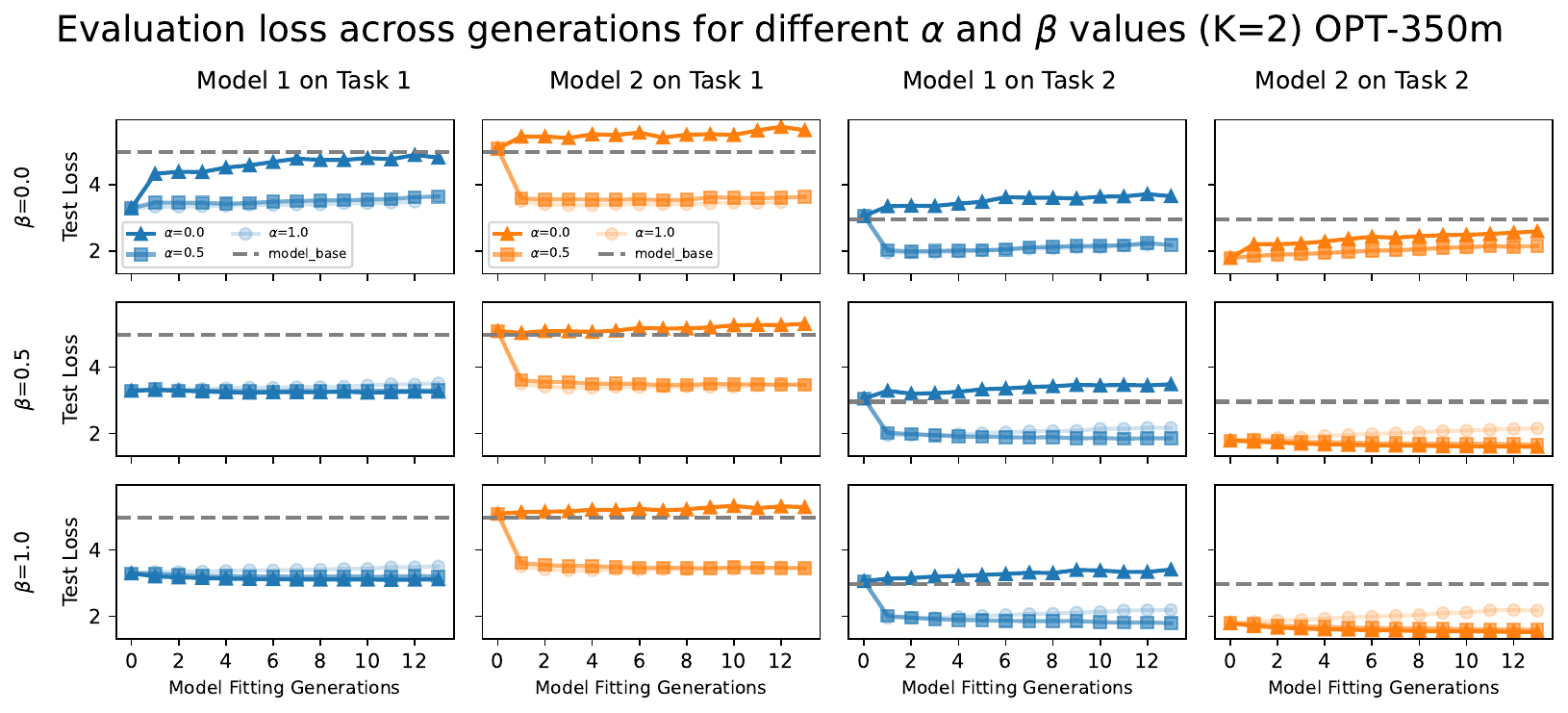}
    \vspace{-0.2cm}
    \caption{\small \textbf{Actual behavior over time for interactions between OPT models ($K=2$) with varying $\alpha,\beta$}.}
    \label{fig:llm_k=2}
    \vspace{-0.2cm}
\end{figure}

\begin{figure}[ht]
    \centering
    \includegraphics[width=0.9\linewidth]{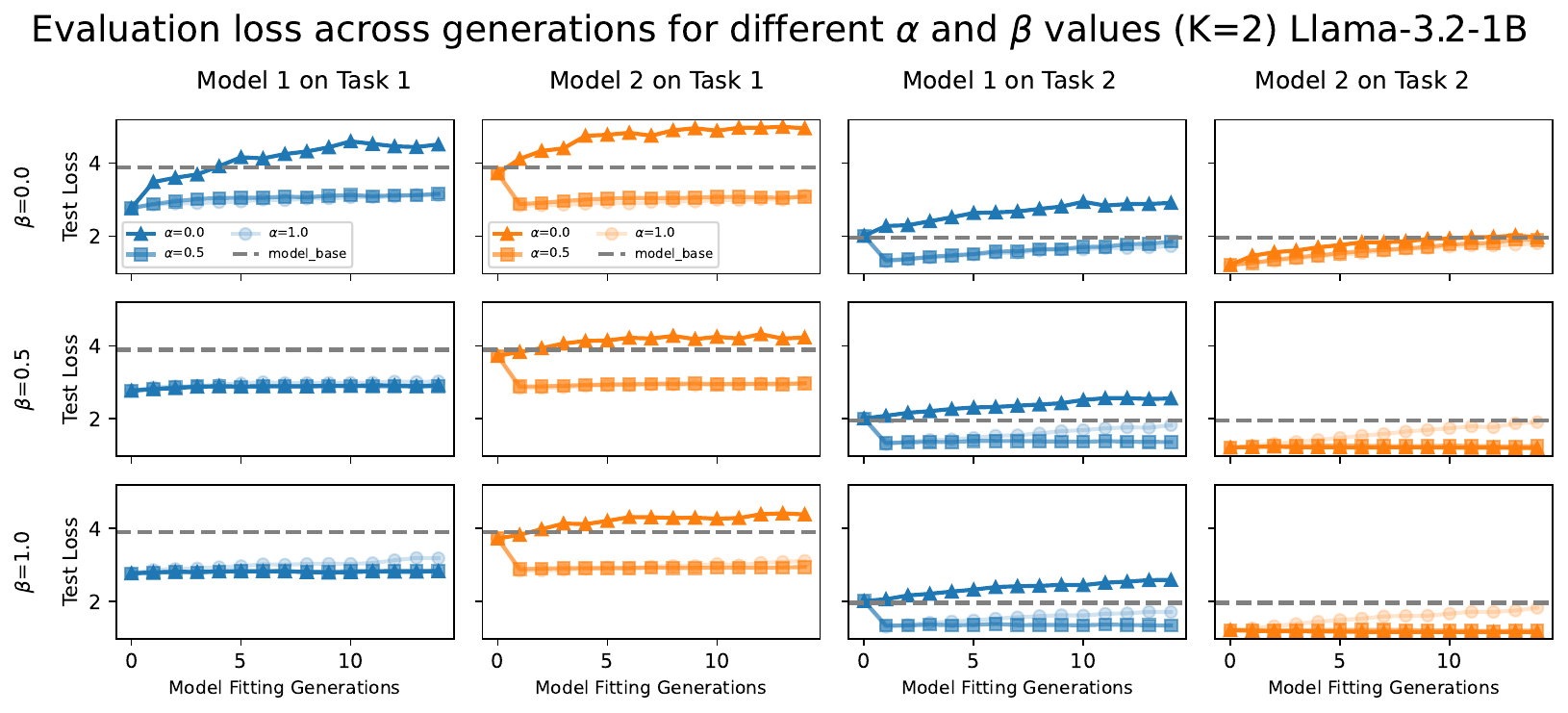}
    \vspace{-0.2cm}
    \caption{\small \textbf{Actual behavior over time for interactions between Llama models ($K=2$) with varying $\alpha,\beta$}.}
    \label{fig:llama_k=2}
    \vspace{-0.2cm}
\end{figure}

\begin{figure}[ht]
    \centering
    \includegraphics[width=0.95\linewidth]{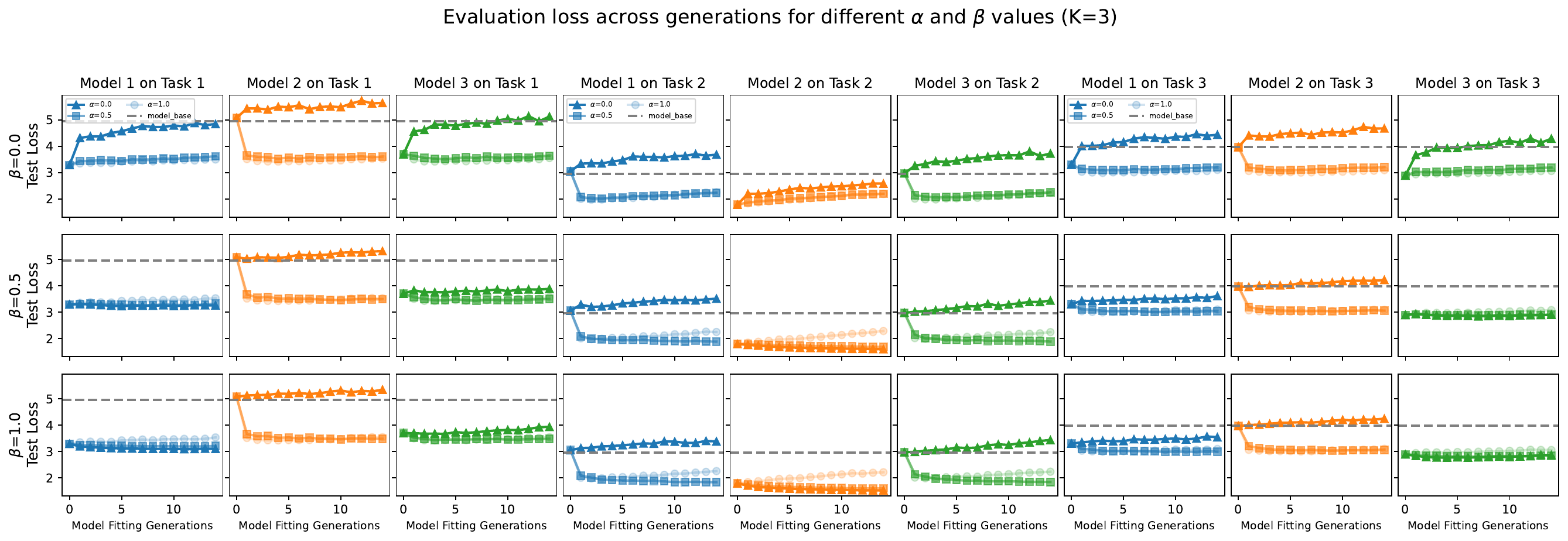}
    \vspace{-0.2cm}
    \caption{\small \textbf{Actual behavior over time for interactions between OPT models ($K=3$) with varying $\alpha,\beta$}.}
    \label{fig:opt_k=3}
    \vspace{-0.2cm}
\end{figure}

Tables~\ref{tab:k2_llama_all} and~\ref{tab:opt_k2_all} show full results for all tested $\beta$, $\alpha$ settings for the $K=2$ setting across both the Llama and OPT model architectures. As stated in the main paper body, the $\beta=0.5$ and $\alpha=0.5$ settings perform best. Figures~\ref{fig:k=2_theory},~\ref{fig:llm_k=2}, and~\ref{fig:llama_k=2} show loss values at each generation for (1) predicted theoretical results, (2) OPT $K=2$ models, and (3) Llama $K=2$ models. Finally, Figure~\ref{fig:opt_k=3} shows results for a $K=3$ system of interacting OPT models.

\begin{table}[]
\centering
\caption{\small {\bf Change in loss behavior for $K=2$ interacting LLama models at various $\alpha, \beta$.} \em We show results as $\texttt{initial} \rightarrow \texttt{final}$ prediction loss values for models on their own and the other models' tasks, at $\texttt{initial} = T=0$  and $\texttt{final} = T =15$ generations. For clarity, we colorize loss \increase{increase}, \decrease{decrease}, and \neutral{constancy} ($\Delta \le 0.1$).}
\label{tab:k2_llama_all}
\resizebox{\textwidth}{!}{%
\begin{tabular}{cccc|ccc|ccc}
\toprule
     & \multicolumn{3}{c|}{$\beta=0$} & \multicolumn{3}{c|}{$\beta=0.5$} & \multicolumn{3}{c}{$\beta=1.0$} \\ \midrule
 &
  $\alpha=0$ & $\alpha=0.5$ &  $\alpha=1.0$ &$\alpha=0$ & $\alpha=0.5$ &  $\alpha=1.0$ &  $\alpha=0$ &  $\alpha=0.5$ &   $\alpha=1.0$  \\ \midrule
  
\begin{tabular}[c]{@{}c@{}}Model 1\\ on Task 1\end{tabular}  &    \increase{$2.8 \rightarrow 4.5$}      &  \increase{$2.8 \rightarrow 3.2$}         &     \increase{$2.8 \rightarrow $ 3.1}  &  \neutral{$2.8 \rightarrow 2.9$}     &     \neutral{$2.8 \rightarrow 2.9 $}     &    \neutral{$2.8 \rightarrow 3.0$}    &    \neutral{$2.8 \rightarrow 2.8$}        &     \neutral{$2.8 \rightarrow 2.8$}      &     \increase{$2.8 \rightarrow 3.2$}     \\ \midrule

\begin{tabular}[c]{@{}c@{}}Model 2\\ on Task 2\end{tabular}  &    \increase{$1.2 \rightarrow 2.0$}     &       \increase{ $1.2 \rightarrow 1.9$}       &    \increase{$1.2 \rightarrow 1.8$}     &   \neutral{$1.2 \rightarrow 1.2$} & \neutral{$1.2 \rightarrow 1.3$} & \increase{$1.2 \rightarrow 1.9$}   &    \neutral{$1.2 \rightarrow 1.2$}       &   \neutral{$1.2 \rightarrow 1.2$}        &   \increase{$1.2 \rightarrow 1.8$}       \\ \midrule \midrule

\begin{tabular}[c]{@{}c@{}}Model 1 \\ on Task 2\end{tabular} &    \increase{$2.0 \rightarrow 2.9$}    &      \neutral{$2.0 \rightarrow 1.9$}      &    \decrease{$2.0 \rightarrow 1.7$}      &   \increase{$ 2.0 \rightarrow 2.5$}  & \decrease{$ 2.0 \rightarrow 1.4$} & \decrease{$ 2.0 \rightarrow 1.8$}      &     \increase{$2.0 \rightarrow 2.6$}       &    \decrease{$2.0 \rightarrow 1.3$}        &     \decrease{$2.0 \rightarrow 1.7$}      \\ \midrule

\begin{tabular}[c]{@{}c@{}}Model 2 \\ on Task 1\end{tabular} &   \increase{$3.7 \rightarrow 5.0$} &  \decrease{$3.7 \rightarrow 3.1$}         &    \decrease{$3.7 \rightarrow 3.1$}      &  \increase{$ 3.7 \rightarrow 4.2 $} & \decrease{$ 3.7 \rightarrow 3.0$} & \decrease{$ 3.7 \rightarrow 3.0$}     &     \increase{$3.7 \rightarrow 4.4$}       &    \decrease{$3.7 \rightarrow 2.9$}      &        \decrease{$3.7 \rightarrow 3.1$}  \\ \bottomrule
\end{tabular}%
}

\end{table}

\begin{table}[]
\centering
\caption{\small {\bf Change in loss behavior for $K=2$ interacting OPT models at various $\alpha, \beta$} \em We show results as $\texttt{initial} \rightarrow \texttt{final}$ prediction loss values for models on their own and the other models' tasks, at $\texttt{initial} = T=0$  and $\texttt{final} = T =15$ generations. For clarity, we colorize loss \increase{increase}, \decrease{decrease}, and \neutral{constancy} ($\Delta \le 0.1$).}
\label{tab:opt_k2_all}
\resizebox{\textwidth}{!}{%
\begin{tabular}{cccc|ccc|ccc}
\toprule
     & \multicolumn{3}{c|}{$\beta=0$} & \multicolumn{3}{c|}{$\beta=0.5$} & \multicolumn{3}{c}{$\beta=1.0$} \\ \midrule
 &
  $\alpha=0$ & $\alpha=0.5$ &  $\alpha=1.0$ &$\alpha=0$ & $\alpha=0.5$ &  $\alpha=1.0$ &  $\alpha=0$ &  $\alpha=0.5$ &   $\alpha=1.0$  \\ \midrule
  
\begin{tabular}[c]{@{}c@{}}Model 1\\ on Task 1\end{tabular}  &    \increase{$3.3 \rightarrow 4.8$}      &  \increase{$3.3 \rightarrow 3.6$}         &     \increase{$3.3 \rightarrow 3.5$}  &  \neutral{$3.3 \rightarrow 3.3$} & \neutral{$3.3 \rightarrow 3.3$} & \increase{$3.3 \rightarrow 3.5$}   &    \decrease{$3.3 \rightarrow 3.1$}        &     \neutral{$3.3 \rightarrow 3.2$}      &     \increase{$3.3 \rightarrow 3.5$}     \\ \midrule

\begin{tabular}[c]{@{}c@{}}Model 2\\ on Task 2\end{tabular}  &    \increase{$1.8 \rightarrow 2.6$}     &       \increase{ $1.8 \rightarrow 2.2$}       &    \increase{$1.8 \rightarrow 2.1$}     &   \neutral{$1.8 \rightarrow 1.7$} & \neutral{$1.8 \rightarrow 1.7$} & \increase{$1.8 \rightarrow 2.2$}  &    \decrease{$1.8 \rightarrow 1.5$}       &   \neutral{$1.8 \rightarrow 1.8$}        &   \increase{$1.8 \rightarrow 2.2$}       \\ \midrule \midrule

\begin{tabular}[c]{@{}c@{}}Model 1 \\ on Task 2\end{tabular} &    \increase{$3.1 \rightarrow 3.7$}    &      \decrease{$3.1 \rightarrow 2.2$}      &    \decrease{$3.1 \rightarrow 2.2$}      &  \increase{$ 3.1 \rightarrow 3.5$} & \decrease{$3.1 \rightarrow 1.8$} & \decrease{$ 3.1 \rightarrow 2.2$}    &     \increase{$3.1 \rightarrow 3.4$}       &    \decrease{$3.1 \rightarrow 1.8$}        &     \decrease{$3.1 \rightarrow 2.2$}      \\ \midrule

\begin{tabular}[c]{@{}c@{}}Model 2 \\ on Task 1\end{tabular} &   \increase{$5.1 \rightarrow 5.6$} &  \decrease{$5.1 \rightarrow 3.6$}         &    \decrease{$5.1 \rightarrow 3.5$}      &  \neutral{$ 5.1 \rightarrow 5.1 $} & \decrease{$5.1 \rightarrow 3.5$} & \decrease{$ 5.1 \rightarrow 3.5 $}    &     \increase{$5.1 \rightarrow 5.4$}       &    \decrease{$5.1 \rightarrow 3.5$}      &        \decrease{$5.1 \rightarrow 3.4$}  \\ \bottomrule     
\end{tabular}%
}

\end{table}

\section{Code for experiments}

Code to generate the theory and experimental figures shown in the main paper body can be found at: \url{https://anonymous.4open.science/r/multi-model-798E}.

\section{Proofs for results in Section~\ref{sec:theory}}

Before diving into the proofs, we recall the workflow setup and provide some preliminary results. The initial data are represented by matrix-vector pairs for the private data  $(\tilde{X}_k, \tilde{y}_k), \dots ,(\tilde{X}_K,\tilde{y}_K)$ and for the public data $(X_*, y_*)$.  At each generation $t$, the data $(X_{tk}, y_{tk})$ produced by entity $k$ is given by
\begin{align}
y_{tk} = X_{tk} \hat{\theta}_{t-1,k} + w_{tk} 
\end{align}
where $\hat{\theta}_{t-1,k} \in \bbR^d$ is the most recent parameter estimate and  $w_{tk} \sim \normal(0, \sigma^2 \Id)$ is Gaussian noise that is independent across entities and generations. To represent the notation compactly, we use the embedding
 \begin{align}
\tilde{\bX} = \begin{bsmallmatrix}\tilde{X}_1 \\ & \ddots \\ && \tilde{X}_K \end{bsmallmatrix} , \quad \by_0  =  \begin{bsmallmatrix}\tilde{y}_1 \\  \vdots \\  \tilde{y}_K   \end{bsmallmatrix}  \quad \bX_t = \begin{bsmallmatrix}X_{t1} \\ & \ddots \\ && X_{tK} \end{bsmallmatrix} , \quad \by_t  =  \begin{bsmallmatrix}y_{tk} \\  \vdots \\  y_{tK} \end{bsmallmatrix} , \quad \bw_t  =  \begin{bsmallmatrix}w_{t1} \\  \vdots \\  w_{tK}   \end{bsmallmatrix} ,
\end{align}

\paragraph{Linear dynamical system} 
According to the workflow, each parameter estimate is obtained as the minimizer in $\theta \in \bbR^d$ of the empirical loss: 
\begin{align}
\bar{\alpha}_t \beta_t \|\tilde{y}_{k}  - \tilde{X}_{k} \theta\|^2  +   \bar{\alpha}_t \bar{\beta}_t  \| y_*   - X_* \theta \|^2  +    \frac{\alpha_t}{K} \sum_{j=1}^K \|y_{tj} - X_{tj} \theta\|^2 ,  
 \end{align}
where $\alpha_0 \equiv 0$ and so only the first two terms are present at initialization. 
The minimum norm solution is given in closed form by 
\begin{align}
\hat{\theta}_{tk} & = \Big(\bar{\alpha}_t \beta_t  \tilde{S}_{k} +    \bar{\alpha}_t \bar{\beta}_t S_*^\top +  \frac{\alpha_t}{K}  \sum_{j=1}^k S_{tk}  \Big)^+ \Big( \bar{\alpha}_t \beta_t \tilde{X}_{k}^\top \tilde{y}_{t}  +  \bar{\alpha}_t \bar{\beta}_t   X_*^\top y_* +  \frac{\alpha_t}{K}  \sum_{j=1}^k X_{tj}^\top y_{tj} \Big) 
\end{align}
where $\tilde{S}_k = \tilde{X}_{t}^\top \tilde{X}_k$, $S_* = X_*X^\top $, $S_{tk} = X_{tk}^\top X_{tk}$, and $(\cdot)^+$ denotes the Moore-Penrose pseudoinverse. Stacking the estimates into a vector $\hat{\btheta} = \gvec(\hat{\theta}_1 ,\dots, \hat{\theta}_K)$ and using the identity $A^\top = A^\top A A^+$, we can express all estimate updates simultaneously as
\begin{align}
\hat{\btheta}_{t} & =\bG_t^+  \Big( \bar{\alpha}_t \beta_t \tilde{\bS}  \tilde{\bX}^+ \tilde{\by}  +  \bar{\alpha}_t \bar{\beta}_t ( \one_K \otimes S_* )X_*^+  y_* +  \frac{\alpha_t}{K} ( \one_K \otimes   \sum_{j=1}^k S_{tj} X_{tj}^+ y_{tj} ) \Big),  \label{eq:hatbtheta_a} 
\end{align}
where $\one_K$ denotes the $K \times 1$ vector of ones and $\bG_t$ is the block diagonal matrix given by 
\begin{align}
\bG_t \coloneqq \bar{\alpha}_t \beta_t  \tilde{\bS} +    \bar{\alpha}_t \bar{\beta}_t  (\Id_K \otimes S_*)  +  \alpha_t( \Id_K \otimes \underline{\bS}_t), \qquad \underline{\bS}_t = \frac{1}{K} \sum_{k=1}^K S_{kt}. 
\end{align}
Defining the orthogonal projection matrix $\Pi = \frac{1}{K} \one_K \one_K^\top \otimes \Id_d$ we can write 
\begin{align}
\frac{1}{K} ( \one_K \otimes   \sum_{j=1}^k S_{tj} X_{tj}^+ y_{tj} ) & = \Pi \bS_t \bX_t^+ \by_t . 
\end{align}
Introducing the matrices  $ \bP_t \coloneqq \bar{\alpha}_t \bG^+_t   \begin{bmatrix}  \beta_t  \tilde{\bS}   &  \bar{\beta}_t  (\one_K \otimes S_*)  \end{bmatrix}$ and $ \bQ_t \coloneqq \alpha_t \bG^+_t \Pi \bS_t$, can express \eqref{eq:hatbtheta_a} as 
\begin{align}
\hat{\btheta}_t & =  \bP_t  \begin{bmatrix}  \tilde{\bX}^+ \tilde{\by} \\ X_*^+ y_*\end{bmatrix}    + \bQ_t  \bX_t^+ \by_t . 
\end{align}
Using  $\by_t = \bX_t \btheta_{t-1}  + \bw_t$ and $\bQ_t \bX_t^+ \bX_t = \bQ_t$, we obtain
\begin{align}
\hat{\btheta}_t & =  \bP_t  \begin{bmatrix}  \tilde{\bX}^+ \tilde{\by} \\ X_*^+ y_*\end{bmatrix}   +    \bQ_t \hat{\btheta}_{t-1} + \bQ_t  \bX_t^+  \bw_t  \label{eq:hatbtheta_c}. 
\end{align}
This expression shows that the estimates evolve according to a discrete-time linear dynamical system (also known as a  Kalman filter model) with state variable  $\hat{\btheta}_t$.  

\subsection{Proof of Theorem~\ref{thm:MtCt}}
We now derive the distribution of $\hat{\btheta}_t$ conditional on the initial data $D_0$ given by  $(\tilde{\bX}, \tilde{\by})$ and $(X_*, y_*)$.  Specifically, we show that the estimates are Gaussian with mean and variance
\begin{align}
  \ex{ \hat{\btheta}_t  \mid D_0 } = \bM_t   
  \begin{bmatrix} \tilde{\bX}^+ \tilde{\by}  \\   X_*^+ y_*  \end{bmatrix} , \qquad 
\cov( \hat{\btheta}_t  \mid  D_0 )  = \bC_t 
\end{align}
where the matrices $\bM_t$ and $\bC_t$ are defined recursively  with $\bM_0 = \bP_0$ and $\bC_0 = \bm{0}_{Kd \times Kd}$ and 
\begin{alignat}{3}
    \bM_t& = \bP_t + \bQ_t \bM_{t-1}, & \qquad   \bC_t &= \bQ_t (\sigma^2 \bS_t^+ +\bC_{t-1} )\bQ_{t}, \qquad t \ge 1.
\end{alignat}

The proof is by mathematical induction.  For the base case $t =0$ we invoke \eqref{eq:hatbtheta_c} along with $\bQ_0 = 0$ to see that  $\hat{\btheta}_0$ is a deterministic function of the initial data with $\bM_0 = \bP_0$ and $\bC_0 = \bm{0}_{Kd \times Kd}$.

For the inductive case, assume that the stated distribution holds up to generation $t-1$. From the definition of the workflow, \eqref{eq:hatbtheta_c}  holds with $\bw_t \sim \normal(0, \sigma^2\Id)$ independent of everything else. 
Thus   $\hat{\btheta}_t$ is Gaussian with mean
\begin{align}
\ex{ \hat{\btheta}_t \mid D_0 }
=  \bP_t  \begin{bmatrix} \tilde{\bX}^+ \tilde{\by}  \\   X_*^+ y_*  \end{bmatrix} +    \bQ_t \ex{ \hat{\btheta}_{t-1} \mid D_0}   = \underbrace{(\bP_t + \bQ_t \bM_{t-1})}_{\bM_t}  \begin{bmatrix} \tilde{\bX}^+ \tilde{\by}  \\   X_*^+ y_*  \end{bmatrix}
\end{align}
and covariance 
\begin{align}
\cov( \hat{\btheta}_t \mid D_0 ) 
& = \cov(  \bQ_t \hat{\btheta}_{t-1}  \mid D_0 ) + \cov(  \bQ_t \bX_t^+  \bw)  = \underbrace{\bQ_t C_{t-1} \bQ_t  +\sigma^2 \bQ_t \bS_t^+ \bQ_t}_{\bC_t}.
\end{align}
This concludes the proof of Theorem~\ref{thm:MtCt}. \qed

\subsection{Proof of Theorem~\ref{thm:bias_mse}}

Under the assumptions of the theorem, we have that
\begin{align}
\begin{bmatrix} \tilde{\bX}^+ \tilde{\by}  \\   X_*^+  y_*  \end{bmatrix}  \sim \normal\left(   \begin{bmatrix} \tilde{\bS} \tilde{\bS}^+ & 0   \\ 0 & S_* S_*^+ \end{bmatrix} ( \one_{K+1} \otimes  \theta),  \sigma^2 \begin{bmatrix} \tilde{\bS}^+ & 0 \\0 &  S^+_* \end{bmatrix}    \right).  \label{eq:ols_dist} 
\end{align}
The goal for this proof is to verify that if $\bG_1, \dots, \bG_t$ are full rank then
\begin{align}
\ex{ \hat{\btheta}_t  } = \left( \Id -  \bQ_t \cdots \bQ_1(\Id -  \bG_0 \bG_0^+)  \right) (\one_K \otimes \theta)
,\quad \cov( \hat{\btheta}_t) = \bM_t \begin{bmatrix}  \tilde{\bS}^+ & 0 \\ 0 & S_*^+ \end{bmatrix} \bM_t^\top + \bC_t.
\end{align}

We proceed by mathematical induction. Consider the case $t =0$. 
By \eqref{eq:hatbtheta_c} along with $\bQ_0 = 0$, the mean is 
\begin{align}
\ex{ \btheta_0} = \bP_t  \begin{bmatrix} \tilde{\bS} \tilde{\bS}^+ & 0   \\ 0 & S_* S_*^+ \end{bmatrix} ( \one_{K+1} \otimes  \theta) = \bG_0 \bG_0^+ (\one_K \otimes \theta) .
\end{align}
Likewise, recalling that $\bM_0 = \bP_0$, the variance is 
\begin{align}
 \cov(  \btheta_0)  = \bM_0 \cov\left( \begin{bmatrix} \tilde{\bX}^+ \tilde{\by}  \\   X_*^+  y_*  \end{bmatrix}   \right) \bM_0^\top =    \sigma^2 \bM_0  \begin{bmatrix} \tilde{\bS}^+ & 0 \\0 &  S^+_* \end{bmatrix}    \bM_0^\top. 
\end{align}

Next, suppose that $\bG_1, \dots, \bG_{t}$ are full rank and the stated distribution holds up to time $t-1$.   By Theorem~\ref{thm:MtCt} we know that $\hat{\btheta}_t$ is Gaussian and so all that remains is to verify the given expressions for the mean and covariance.  By the linearity of expectation  and \eqref{eq:hatbtheta_c}, the mean satisfies 
\begin{align}
\ex{ \hat{\btheta}_t} 
& = \bP_t \begin{bmatrix} \ex{  \tilde{\bX}^+ \tilde{\by} }   \\   \ex{ X_*^+  y_* }   \end{bmatrix}    + \bQ_t \ex{ \hat{\btheta}_{t-1}} \\
& = \bP_t  (\Id_{K+1} \otimes \theta)    + \bQ_t (\Id_K \otimes \theta)  -   \bQ_t   \bQ_{t-1} \cdots \bQ_1(\Id -  \bG_0 \bG_0^+)  (\one_K \otimes \theta),
\end{align}
where the last follows from the inductive assumption applied to $\ex{\hat{\btheta}_{t-1}}$.  Moreover, from the definitions of $\bP_t$ and $\bQ_t$, we have 
\begin{align}
 \bP_t  (\Id_{K+1} \otimes \theta)    + \bQ_t (\Id_K \otimes \theta)   & =  \bG^+_t \Big (   \bar{\alpha}_t \beta_t \tilde{\bS}  +  \bar{\alpha}_t \bar{\beta}_t  (\Id_K \otimes S_*) )  + \alpha_t \Pi \bS_t  \Big) (\Id_K \otimes \theta) \\
 & =  \bG^+_t \bG_t (\Id_K \otimes \theta) \\
  & =\Id_K \otimes \theta,
\end{align}
where the second line follows from the identity $\Pi \bS_t (\one_K \otimes \Id_d) = (\Id_K \otimes \underline{\bS}_t) (\one_K \otimes \Id_d) $
and the last line holds because $\bG_t$ is full rank. Combining the above displays gives the desired expression for the mean. The expression for the variance follows directly from \eqref{eq:ols_dist} and Theorem~\ref{thm:MtCt}. \qed

\subsection{Proof of Theorem~\ref{thm:asymp_var}}

If the spectral radius of $\bQ$ is strictly less than one, then  $\bQ^t \to \bm{0}$ as $t \to \infty$, and the Neumann series in \eqref{eq:MtCt_alt} converge to the well-defined limits in \eqref{eq:MC}. These limits can also be seen as the (necessarily unique) solutions to the fixed point equations
\begin{align}
\bM = \bP + \bM \bQ, \qquad \bC = \bQ ( \bC + \sigma^2 \bS^+ ) \bQ^\top, 
\end{align}
where the expression for the covariance is known as the discrete time Lyapunov equation.  Combining these convergence results with Theorem~\ref{thm:bias_mse} completes the proof. \qed.

\subsection{Proof of Lemma~\ref{lem:convergence_cond}}
If $\alpha =0$ or if $\bS = \bm{0}$ then $\bQ = \bm{0}$ and so the stated result holds. Henceforth, we assume  $0 < \alpha < 1$ and $\bS$ is nonzero. Suppose that  $\gamma \bS = \lambda \tilde{\bS}  +(1-\lambda)  (\Id_K \otimes S_*)$ for some $0 < \beta  \le \lambda \le 1$ and $\gamma > 0$. Then,
\begin{align}
\bG & =  \bar{\alpha} \beta \tilde{\bS}  + \bar{\alpha} \bar{\beta} ( \Id_K \otimes S_* ) + \alpha ( \Id_K \otimes \underline{\bS} )\\
& =  \frac{\bar{\alpha} \beta}{ \lambda} ( \gamma  \bS  - (1- \lambda)  (\Id_K \otimes S_*) )   + \bar{\alpha} \bar{\beta} ( \Id_K \otimes S_* ) + \alpha ( \Id_K \otimes \underline{\bS} )\\
& =  \frac{\bar{\alpha} \beta \gamma }{ \lambda}  \bS   + \bar{\alpha} \left(\frac{ \lambda -  \beta}{ \lambda} \right)  ( \Id_K \otimes S_* ) + \alpha ( \Id_K \otimes \underline{\bS}).
\end{align}
Hence, 
\begin{align}
\bQ = \big( \delta \bS  + \Id_K \otimes \Delta  \big)^{+} \Pi \bS, \qquad \delta = \frac{ \bar{\alpha} \beta \gamma }{  \alpha \lambda} , \qquad \Delta =  \frac{ \lambda - \beta}{  \alpha \lambda } S_* + \underline{\bS}. 
\end{align}
To proceed, observe that each diagonal block of $\bS = (S_1,  \dots, S_K)$ lies in the span of $\underline{\bS}= \frac{1}{K} \sum_{k=1}^K S_k$, and thus  $\bS$ lies in the span of $\Id_K \otimes \Delta$. Accordingly, we can write 
\begin{align}
\bS  \big( \delta \bS  + \Id_K \otimes \Delta  \big)^{+} & = ( \Id_K \otimes \Delta^{1/2} )   \bR  \big( \delta \bR  + \Id  \big)^{-1} ( \Id_K \otimes \Delta^{+/2} ), 
\end{align}
where $(\cdot)^{1/2}$ denote the symmetric positive semidefinite square root of a positive semidefinite and  $\bR \coloneqq ( \Id_K \otimes \Delta^{+/2} ) \bS ( \Id_K \otimes \Delta^{+/2} )$. To bound the spectral radius, denoted by $\rho(\cdot)$, we use that fact that the eigenvalues of $A B$ and $BA$ are the same for any square matrices $A$ and $B$ along with that fact that $\Id_K \otimes \Delta$ commutes with $\Pi$ to write 
\begin{align}
\rho(\bQ) & = \rho\left( \big( \delta \bS  + \Id_K \otimes \Delta  \big)^{+} \Pi  \bS \right) \\
& = \rho\left( \bS \big( \delta \bS  + \Id_K \otimes \Delta  \big)^{+} \Pi \right) \\
& = \rho\left(  ( \Id_K \otimes \Delta^{1/2} )   \bR  \big( \delta \bR  + \Id  \big)^{-1} ( \Id_K \otimes \Delta^{+/2} )  \Pi \right) \\
& = \rho\left( \Pi  \bR \big( \delta \bR  + \Id \big)^{-1}   \Pi \right)  \\
& =\|   \Pi \bR \big( \delta \bR  + \Id \big)^{-1}   \Pi  \|,
\end{align}
where $\|\cdot\|$ denotes the operator norm and the last equality holds because $\Pi \bR \big( \delta \bR  + \Id \big)^{-1}   \Pi$ is symmetric positive semidefinite. Letting $\eps > 0$ denote the smallest nonzero singular value of $\bR$, we have 
\begin{align}
(1 + \eps \delta) \rho(\bQ) & \le \|   \Pi \bR   \Pi  \|   = \| (\Id_K \otimes \Delta^{+/2} )    \Pi \bS \Pi (\Id_K \otimes \Delta^{+/2} )   \|\\
&   = \| (\Id_K \otimes \Delta^{+/2} )    (\Id_K \otimes \underline{\bS})  (\Id_K \otimes \Delta^{+/2} )   \| \le 1.
\end{align}
This verifies that $\rho(\bQ)$ is strictly less than one. \qed

\end{document}